\pdfoutput=1
\documentclass[11pt]{article}

\usepackage{ACL2023}

\usepackage{times}
\usepackage{latexsym}

\usepackage{verbatim}
\usepackage[T1]{fontenc}
\usepackage[utf8]{inputenc}

\usepackage{microtype}

\usepackage{inconsolata}
\usepackage{enumitem}
\usepackage{amssymb}
\usepackage{amsmath}
\usepackage{algorithm}
\usepackage{siunitx} 
\usepackage{todonotes}
\usepackage{makecell}
\usepackage{algorithmic}
\usepackage{booktabs}
\usepackage{threeparttable} % Allows for adding notes and footnotes in tables
\usepackage{multirow}
\usepackage{enumitem}
\usepackage{placeins}
\usepackage{geometry}
\usepackage{array}
\usepackage{colortbl} % Required for \cellcolor
\usepackage{tcolorbox}
\usepackage{fontawesome5} 

\title{From Annotation to Adaptation: Metrics, Synthetic Data, and Aspect Extraction for Aspect-Based Sentiment Analysis with Large Language Models}

\author{
  \textbf{Nikita Neveditsin\textsuperscript{1}},
  \textbf{Pawan Lingras\textsuperscript{1}},
  \textbf{Vijay Mago\textsuperscript{2}}
\\
\textsuperscript{1}Saint Mary's University, Halifax, Canada \\
\textsuperscript{2}York University, Toronto, Canada
\\
}

\definecolor{palered}{RGB}{255, 204, 204}
\definecolor{palegreen}{RGB}{204, 255, 204}

\begin{document}
\maketitle
\begin{abstract}
This study examines the performance of Large Language Models (LLMs) in Aspect-Based Sentiment Analysis (ABSA), with a focus on implicit aspect extraction in a novel domain. Using a synthetic sports feedback dataset, we evaluate open-weight LLMs' ability to extract aspect-polarity pairs and propose a metric to facilitate the evaluation of aspect extraction with generative models. Our findings highlight both the potential and limitations of LLMs in the ABSA task.

\end{abstract}

\setlength{\textfloatsep}{0pt}
\captionsetup[table]{skip=0pt}
\setlength{\tabcolsep}{2pt}
\renewcommand{\arraystretch}{1}
\begin{table*}[h!]
\scriptsize 
\centering
\begin{threeparttable}
\caption{ABSA Subtasks.}
\label{tab:absa-tasks}
\begin{tabular}{@{}p{0.7\textwidth}p{0.3\textwidth}@{}}
\toprule
\textbf{Subtask Names} & \textbf{Extracted Entity} \\ 
\midrule
Aspect Extraction
\cite{liu2022sentiment},
Opinion Target Expression Extraction 
\cite{pontiki-etal-2015-semeval}, 
Aspect Term Extraction (ATE)
\cite{pontiki-etal-2014-semeval, scaria2023instructabsa}  & 
Aspect \space (e.g., "restaurant atmosphere", "technical support")  %\tnote{1}
\\ 
\midrule
Aspect Sentiment Classification 
\cite{liu2022sentiment},
Sentiment Polarity Classification
\cite{pontiki-etal-2015-semeval}, 
Aspect Term Polarity Classification 
\cite{pontiki-etal-2014-semeval} & 
Polarity (e.g., "positive", "negative", "neutral") \\ 
\midrule
Aspect Category Detection 
\cite{pontiki-etal-2014-semeval} & 
Category (e.g., "food") \\  %\tnote{2}
\midrule
Opinion Term Extraction 
\cite{WOS:001089176900009} & 
Opinion Phrase (e.g, "could be better") \\ 
\bottomrule
\end{tabular}
%\begin{tablenotes}%[para, flushleft]
%\setlength\itemsep{0pt}
%\item[1] Aspects can be explicit or implicit and can be pre-defined or extracted from a document.
%\item[2] Aspect categories can be pre-defined or inferred from a set of documents.
%\end{tablenotes}
\end{threeparttable}
\vspace{-10pt} 
\end{table*}

\section{Introduction}
\setlength{\parskip}{1pt}

ABSA is a nuanced form of sentiment analysis that focuses on identifying sentiments related to specific aspects within a text \cite{pontiki-etal-2014-semeval}. Researchers have decomposed ABSA into various subtasks, such as aspect extraction, sentiment classification, aspect category detection, and opinion term extraction, each contributing to a comprehensive understanding of the problem. Table \ref{tab:absa-tasks} summarizes these subtasks as discussed in the literature. Combining these tasks allows the extraction of ABSA-related entities in the form of tuples, triples, or quadruples from sentences or documents, resulting in a wide range of compound ABSA solutions. 

LLMs with their in-context learning (ICL) capabilities \cite{DBLP:journals/corr/abs-2005-14165} and parameter-efficient fine-tuning methods, such as Low-Rank Adaptation (LoRA) with quantization \cite{dettmers2024qlora, hu2021lora}, offer straightforward yet effective approaches for complex ABSA tasks. These approaches facilitate the extraction of \textit{implicit aspects}, which are aspects that are not explicitly stated in the text but can be inferred based on context, sentiment, or background knowledge.

This study examines the performance of LLMs in extracting aspect-polarity pairs within the under-explored and unanticipated domain of sports feedback. This domain poses unique challenges for ABSA due to its reliance on implicit references and domain-specific terminology. By evaluating LLMs in this context, we provide critical insights into their capacity to adapt to novel data.

Moreover, recognizing the linguistic variability involved in expressing implicit aspects, we propose an evaluation metric that calculates precision and recall while accounting for this variability in settings with a high prevalence of implicit aspects. We also demonstrate the broader applicability of this metric, showing its utility in assessing generative LLMs on classic ABSA datasets. Finally, we explore various strategies for adapting LLMs to domain-specific datasets, highlighting key challenges and offering insights for future research.

\section{Related Work}
\subsection*{Aspect-Based Sentiment Analysis}

Traditional approaches to ABSA, extensively reviewed in the literature \cite{nazir2020issues, 10.1145/3503044, WOS:001089176900009}, primarily utilize bidirectional encoders \cite{dos2021multi, zhang-etal-2023-span}, recurrent networks \cite{xu2020target}, graph networks \cite{huang, WU2022107736, wang-etal-2024-dagcn}, sequence-to-sequence models \cite{ma2019exploring}, and ensembles of models \cite{pyabsa}. Various techniques have recently been proposed to improve accuracy, precision, and recall in ABSA-related tasks, for example, context denoising \cite{tian-etal-2024-aspect}, abstract meaning representation \cite{ma-etal-2023-amr}, and global semantic features \cite{zhou-etal-2024-gcnet}. These methods have achieved robust results in within-domain explicit aspect extraction and polarity classification \cite{MengABSA, MESKELE2020102211, wang2020relational}. 

Recent studies have investigated the ability of LLMs to perform ABSA tasks on both traditional \cite{smid-etal-2024-llama} and more complex datasets \cite{deng2023llms, krugmann2024sentiment}, highlighting the potential of generative models in key ABSA subtasks \cite{kheiri2023sentimentgpt, scaria2023instructabsa, yang2024faima}. Nevertheless, challenges persist in effectively capturing implicit aspects, particularly in low-resource domains, where difficulties in data collection and annotation further exacerbate the problem \cite{TUBISHAT2018545, wankhade2022survey, cai2021aspect, WOS:001089176900009}.

\subsection*{Data Creation and Annotation for ABSA}

Advancing ABSA research can benefit from quality datasets. Recent work by \citet{oats} demonstrated that human annotation of ABSA datasets involving implicit aspects is challenging and laborious. Generative LLMs have been successfully utilized to create and annotate synthetic datasets, leveraging their capacity to generate creative and contextually rich text \cite{meyer2022we, bao-etal-2023-synthetic, eldan2023tinystories, Mirowski}. Although LLMs may not always match human annotators in accuracy, studies have shown that their annotations can be valuable, particularly when combined with human expertise \cite{pmlr-v225-goel23a, gray2023can, pmlr-v239-mohta23a, gpt4annot, liyanage2024gpt}.

Moreover, leveraging synthetic data has been explored to enhance the performance of downstream models in various NLP tasks, including ABSA \cite{kramchaninova2022synthetic, yu2023cross, deng2023llms, wang2024refining}.

\section{Datasets}

\subsection{Novel dataset}
We introduce a novel dataset of artificially generated feedback from volunteers at sports event, a domain not yet represented in existing ABSA datasets. This domain poses unique challenges due to its specific terminology and the abundance of implicit aspects. The dataset facilitates an out-of-domain evaluation of the ABSA capabilities of open-weight LLMs against baseline solutions. Notably, at least 35\% of its content comprises implicit aspects\footnote{Aspects that do not exactly match any part of a document.}. Additionally, the dataset's domain specificity provides an opportunity to test the generalization capabilities of ABSA solutions beyond their usual training contexts, contributing to a deeper understanding of their real-world applicability. 

We chose two state-of-the-art models\footnote{As of March 2024, when the dataset was generated and annotated} for dataset generation: GPT-4 and Gemini 1.0 Ultra. The novel dataset comprises 480 documents, with an average of 222 characters per document. Most of the dataset (75\%) was generated using GPT-4, acknowledging its superior reported results for major benchmarks such as MMLU \cite{openai2023gpt4}. Additionally, we employed Gemini 1.0 Ultra to generate 25\% of the dataset, introducing some diversity of content. Appendix~\ref{sec:promts} provides examples of prompts and generated text, illustrating the models' ability to produce mixed-emotion and diverse style feedback.

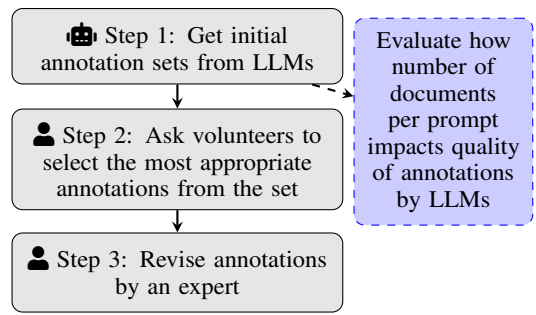
\begin{figure}[h!]
    \centering
    \begin{tikzpicture}[node distance=1.5cm]
    \footnotesize

    % Define block styles
    \tikzstyle{block} = [rectangle, rounded corners, minimum width=4cm, minimum height=1cm, text centered, draw=black, fill=gray!20, text width=4cm, inner sep=5pt]
    \tikzstyle{block_blue} = [rectangle, rounded corners, minimum width=2cm, minimum height=1cm, text centered, draw=blue, fill=blue!20, text width=2cm, inner sep=5pt, dashed]
    \tikzstyle{arrow} = [thick,->,>=stealth]
    \tikzstyle{dashed_arrow} = [thick,->,>=stealth, dashed]

    % Nodes with icons
    \node (step1) [block] {\faRobot{} Step 1: Get initial \\ annotation sets from LLMs};
    \node (bluebox) [block_blue, right of=step1, xshift=2cm, yshift=-1cm] {Evaluate how number of documents per prompt impacts quality of annotations by LLMs};

    % Lower nodes with icons
    \node (step2) [block, below of=step1] {\faUser{} Step 2: Ask volunteers to \\ select the most appropriate \\ annotations from the set};
    \node (step3) [block, below of=step2] {\faUser{} Step 3: Revise annotations \\ by an expert};

    % Arrows
    \draw [arrow] (step1) -- (step2);
    \draw [arrow] (step2) -- (step3);
    \draw [dashed_arrow] (step1) -- (bluebox);

    \end{tikzpicture}
    \caption{Workflow of the Annotation Process.}
    \label{fig:annotation}
\end{figure}

The dataset annotation process, illustrated in Figure~\ref{fig:annotation}, involved three steps, integrating both LLMs and human annotators. First, LLMs generated initial annotation drafts to alleviate the cognitive and time burden on the expert. Next, volunteers selected the better draft from two options. Finally, the expert revised and refined the selected draft.

Appendix~\ref{sec:annotproc} provides a detailed description of the dataset annotation process. We make the dataset and the prompts used for its generation publicly available\footnote{\url{https://github.com/neveditsin/absa-sport}} and publish the Datasheet for the dataset, as proposed by \citet{gebru2021datasheets}, in the same repository.

\subsection{Existing Datasets}
For this study, we specifically selected existing datasets that are well-suited for the joint task of detection of aspects and the classification of their polarities. While numerous other datasets are available \cite{chebolu2023survey, WOS:001089176900009}, we restricted our choices to those documented in published, peer-reviewed papers to ensure higher annotation quality. Table \ref{tab:absa_dss} summarizes these datasets and includes statistics for the novel dataset we introduce in this paper in the last row. Appendix~\ref{sec:stats} provides additional characteristics of the datasets.

%\vspace{-8pt}
%\captionsetup[table]{skip=0pt}
\begin{table}[h]
\begin{threeparttable}
\caption{Datasets Used for Experiments.} 
\centering
\footnotesize 
\setlength{\tabcolsep}{2pt}
\begin{tabular}{p{5cm}|c|c|c}
 & Train & Test & Implicit \\ 
 & & & Aspects \\ 
\hline
SemEval-14-Laptop
\cite{pontiki-etal-2014-semeval} & 1482 & 422 & 0\% \\
SemEval-14-Restaurant 
\cite{pontiki-etal-2014-semeval} & 2019 & 606 & 0\% \\
MAMS 
\cite{jiang-etal-2019-challenge} & 4297 & 500 & 0\% \\
Twitter \cite{dong2014adaptive}  & 6248 & 692 & 3.5\% \\
\hline
Composite & 14046 & 2220 & 0.88\% \\
\hline
Sports Feedback (Novel) & 96 & 384 & 35\% \\
\end{tabular}
\label{tab:absa_dss}
\end{threeparttable}
%\vspace{-20pt}
\end{table}

%\vspace{-15pt}

%\cite{dong-etal-2014-adaptive}

\section{Metrics}

Automated evaluation of models for the aspect detection subtask faces several challenges. First, documents may contain implicit aspects that do not directly match with individual words. For example, the sentence from our dataset:

%\vspace{-3pt}
\begin{quote}
\textit{I found that some locations had multiple volunteers that didn't appear to be overly busy and could have been useful at other locations where there were shortages.}
\end{quote}
%\vspace{-3pt}

This sentence alludes to the aspect `allocation of volunteers' without explicitly stating it in the text. Moreover, the definition of what constitutes an aspect is often fuzzy: in the cited example, `placement of volunteers' could also be interpreted as a valid aspect.

Second, when LLMs are used for aspect extraction instead of traditional span-based approaches, relying on exact matches to compute metrics such as precision, recall, and F-score without accounting for linguistic variation can be problematic.

\begin{figure}[H]
    \centering
    \resizebox{0.4\textwidth}{!}{
        \begin{tikzpicture}
            % Define colors
            \definecolor{set1color}{RGB}{255,204,153}  % Light orange for S1
            \definecolor{set2color}{RGB}{173,216,230}  % Light blue for S2
            \definecolor{intersectioncolor}{RGB}{144,238,144}  % Light green for intersection
            
            % Draw two sets with transparency
            \filldraw[fill=set1color, fill opacity=0.4, draw=black, thick] (0,0) circle(3cm); % Set S1 with gold color
            \filldraw[fill=set2color, fill opacity=0.4, draw=black, thick] (4,0) circle(3cm); % Set S2 with greenish color
            
            % Labels for sets, adjusted to be above the circles
            \node at (-1,3.5) {$S_g$ (Gold Aspects)};
            \node at (4.5,3.5) {$S_d$ (Detected Aspects)};
            
            % Elements in S1
            \node at (-1.5,1.5) {AC};
            \node at (-2,0.5) {look};
            \node at (-1.5,-0.5) {ambience};
            \node at (-1.5,-1.5) {service};
            
            % Elements in S2
            \node at (4.5,1.5) {air conditioner};
            \node at (5.7,0.4) {appearance};
            \node at (6,-0.5) {dishes};
            \node at (5.5,-1.5) {service};
            \node at (4,-2) {drinks};
            
            % Intersection dots (evenly distributed in the intersection area)
            \node at (2,2.7) {\textbf{$\iota$}};
            
            \filldraw[black] (2,1) circle(2pt);   % Dot for AC-air conditioner pair
            \filldraw[black] (2,-0.125) circle(2pt); % Dot for look-appearance pair
            \filldraw[black] (2,-1.25) circle(2pt);  % Dot for service-service pair
            
            % Edges from S1 to intersection
            \draw[thick] (-1.2,1.5) -- (2,1);        % AC to dot
            \draw[thick] (-1.6,0.4) -- (2,-0.125);    % look to dot
            \draw[thick] (-0.9,-1.5) -- (2,-1.25);   % service to dot
            
            % Edges from S2 to intersection
            \draw[thick] (3.3,1.4) -- (2,1);         % air conditioner to dot
            \draw[thick] (4.8,0.4) -- (2,-0.125);     % appearance to dot
            \draw[thick] (4.9,-1.5) -- (2,-1.25);    % service to dot
            
            % Shading for the intersection
            \begin{scope}
                \clip (0,0) circle(3cm);
                \fill[intersectioncolor, opacity=0.5] (4,0) circle(3cm); % Intersection with light blue color
            \end{scope}
            
        \end{tikzpicture}
    }

    \caption{Intersection $\iota$ of Gold Aspects ($S_g$) and Detected Aspects ($S_d$).}
    \label{fig:intersection}
\end{figure}
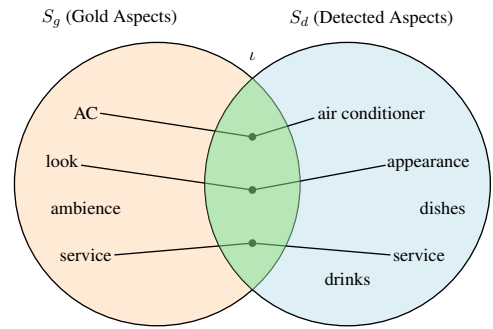

To address these evaluation challenges, we propose a generalized method for assessing precision ($P$) and recall ($R$) inspired by the work of \citet{euzenat2007semantic} on ontology alignment. Specifically, to account for partial matches and linguistic variation between predicted and true aspect sets, we define precision and recall as follows. For a given document, we define $S_d$ as the set of detected aspects and $S_g$ as the set of true (gold) aspects. The function $\iota$, parameterized by a threshold $\theta \in [0, 1]$, returns the set of partial matches between $S_d$ and $S_g$. Figure~\ref{fig:intersection}  illustrates the concept of the intersection $\iota$ between the two sets of aspects.

The threshold $\theta$ serves as a filter for the minimal similarity required between pairs of matching aspects. In the special case where $\theta = 1$, the function $\iota(S_d, S_g)$ reduces to the intersection of the two sets, enforcing exact aspect matches. Conversely, when $0 \leq \theta \ll 1$, it permits the matching of semantically unrelated pairs, making values of $\theta$ close to zero impractical. For the purpose of experiments in this study, we set $\theta = 0.95$. An empirical analysis of the impact of $\theta$ on matching errors in the context of this study is provided in Appendix~\ref{sec:metricanalysis}.

With these definitions, the generalized precision, denoted as $P^{\theta}$, is given by:
\begin{equation}
P^{\theta} = \frac{\lvert \iota(S_{d}, S_{g}, \theta) \lvert}{\lvert S_{d} \lvert}
\end{equation}

Similarly, the generalized recall, denoted as $R^{\theta}$, is formulated as:
\begin{equation}
R^{\theta} = \frac{\lvert \iota(S_d, S_g, \theta) \lvert}{\lvert S_g \lvert}
\end{equation}
%\vspace{-0.3em}

The $F_1^{\theta}$ score, defined as the harmonic mean of precision $P^{\theta}$ and recall $R^{\theta}$, effectively captures the balance between these metrics within this framework.

Algorithm 1 provides the implementation of the function $\iota(S_g, S_d, \theta)$ used in this study. A similarity measure \( \sigma: s_1 \times s_2 \to [0,1] \) quantifies the resemblance between individual elements from the sets, resulting in a similarity matrix with values ranging from 0 to 1. To avoid false positive matches, values below a specified threshold \( \theta \) are set to zero. The similarity matrix is then converted into a cost matrix, and the linear sum assignment problem is solved to determine the optimal pairing of elements between the sets, minimizing the total cost. This procedure yields a set of optimal element pairs, \( \mathcal{I} \).

\begin{algorithm}[H]
\footnotesize
\caption{Algorithm for Finding Intersection $\iota$}
\begin{algorithmic}[1]
\REQUIRE Two finite sets of aspects \( S_g \) and \( S_d \); similarity measure \( \sigma: s_1 \times s_2 \to [0,1] \); similarity threshold \( \theta \)
\ENSURE Optimal pairing set \( \mathcal{I} \) of index pairs \( (i, j) \)
\STATE Initialize similarity matrix \( M \) of size \( |S_g| \times |S_d| \)
\FOR{each \( s_{1i} \in S_g \)}
    \FOR{each \( s_{2j} \in S_d \)}
        \STATE \( M_{ij} \leftarrow \sigma(s_{1i}, s_{2j}) \)
    \ENDFOR
\ENDFOR
\FOR{each element \( M_{ij} \) in \( M \)}
    \IF{\( M_{ij} < \theta \)}
        \STATE \( M_{ij} \leftarrow 0 \)
    \ENDIF
\ENDFOR
\STATE Define cost matrix \( C \) where \( C_{ij} \leftarrow 1 - M_{ij} \)
\STATE Solve the linear sum assignment problem using \( C \) to obtain optimal pairing set \( \mathcal{I} \)
\RETURN \textbf{\( \mathcal{I} \)}
\end{algorithmic}
\end{algorithm}

For this study, we use the algorithm described by \citet{Crouse} to solve the linear sum assignment problem and implement the function $\sigma(s_1, s_2)$ as the scaled cosine similarity between the embeddings of $s_1$ and $s_2$. %, generated using the Sentence-T5-Large model \cite{ni2021sentencet5}.

\section{Models}

We evaluated two open-weight models, Mistal 7B Instruct \cite{jiang2023mistral} and LLaMA-3 8B Instruct \cite{bhatt2024cyberseceval}, against the baseline PyABSA \cite{pyabsa} on the Aspect-Polarity Pair Extraction (ASPE) task. The selection of the open-weight models was motivated by their state-of-the-art performance within the parameter range\footnote{As of July 2024}, ease of deployment, and computational efficiency. Their relatively compact sizes (7–8 billion parameters) allow local deployment without reliance on external computational resources, a material factor for practical applications.

PyABSA is an actively maintained, ensemble-based framework trained on publicly available datasets. It serves as a reliable baseline representing traditional yet robust ABSA methodologies.

For measuring phrase similarity, we selected Sentence-T5 (Large) \cite{ni2021sentencet5}. Despite its smaller size compared to more recent large-scale models, Sentence-T5 demonstrates strong performance on text embedding benchmarks \cite{mteb}, making it well-suited for experiments with limited computational resources.

\section{Evaluation of Open-Weight Models}
% (based on DeBERTa-v2 \cite{he2021deberta})
Our experiments aim to address the following research questions:
\begin{enumerate}[itemsep=3pt, parsep=0pt, topsep=0pt]
    \item Can open-weight LLMs outperform the baseline without fine-tuning?
    \item How do in-context learning examples affect the performance of LLMs on the ASPE task? 
    \item Does fine-tuning on (i) similar data or (ii) data from a different domain with a large fraction of implicit aspects improve the performance of the selected LLMs on the joint task compared to the baseline and non-fine-tuned models?
\end{enumerate}

For the experiments, we organized the datasets from Table~\ref{tab:absa_dss} into two categories: (i) the \textit{Novel dataset}, introduced in this paper, and (ii) the \textit{Composite dataset}, assembled by aggregating the previously published datasets listed in Table~\ref{tab:absa_dss}. For model evaluation, we used the test sets from both datasets: 2,220 samples from the Composite dataset and 384 samples from the Novel dataset. 

For model fine-tuning, we utilized: 
\begin{enumerate}[itemsep=3pt, parsep=0pt, topsep=0pt]
    \item The training portion of the Composite dataset, containing 14,046 samples. 
    \item The training portion of the Novel dataset, consisting of 96 samples. Due to its limited size, we allocated 80\% of the Novel dataset to testing and 20\% to training.
    \item A blended dataset obtained by combining the training portion of the Novel dataset (96 samples) with 96 randomly selected samples from each of the existing datasets listed in Table~\ref{tab:absa_dss}, resulting in a total of 480 samples.
\end{enumerate}

Appendix~\ref{sec:fthyper} provides the complete set of fine-tuning hyperparameters and lists the hardware and software used for the experiments.

For ICL examples, we uniformly sampled documents along with their associated sets of aspect-polarity pairs from the training subset of the respective dataset: when evaluating on the Novel dataset, we sampled from its training subset, and when evaluating on the Composite datasets, we sampled from its training portion. For each polarity in $\mathcal{P} = \{\text{positive}, \text{neutral}, \text{negative}\}$, two documents were selected to ensure compatibility with the model's context window during inference.

Table~\ref{tab:detailed-results} compares the performance of fine-tuned models, generic ICL (using the same predefined prompt with arbitrary examples presented to the models; see Appendix~\ref{sec:genericicl} for reference), ICL with sampling, and a baseline on the aspect extraction subtask. The evaluation employs macro-averaged metrics with a threshold $\theta = 0.95$. This threshold, empirically chosen to accommodate variations in aspect phrasing while minimizing errors, is analyzed in detail in Appendix~\ref{sec:metricanalysis}.

\begin{table}[h]
\centering
\scriptsize % Smaller font size
\setlength{\tabcolsep}{4pt} % Adjust column spacing
\renewcommand{\arraystretch}{1} % Less vertical space between rows
\caption{Experimental Results for Aspect Extraction.}
\label{tab:detailed-results}

\begin{tabular}{@{}l@{\hspace{2pt}}c@{\hspace{5pt}}ccc@{\hspace{5pt}}ccc@{}}
\toprule
\multirow{2}{*}{\textbf{Model}} & \multirow{2}{*}{\textbf{Fine-Tuning / ICL}} & \multicolumn{3}{c}{\textbf{Composite Dataset}} & \multicolumn{3}{c}{\textbf{Novel Dataset}} \\
\cmidrule(lr){3-5} \cmidrule(lr){6-8}
& & \( P^{.95} \) & \( R^{.95} \) & \( F_1^{.95}\) & \( P^{.95} \) & \( R^{.95} \) & \( F_1^{.95} \) \\
\midrule

% No Fine-Tuning
Mistral & Generic ICL & \cellcolor{palered}0.35 & 0.59 & \cellcolor{palered}0.44 &\cellcolor{palered} 0.21 & 0.44 & \cellcolor{palered}0.29 \\
LLaMA-3 & Generic ICL & 0.49 & 0.59 & 0.53 & 0.33 & 0.51 & 0.40 \\

% Sampled ICL
Mistral & ICL with sampling & 0.68 & 0.63 & 0.65 & 0.52 & 0.50 & 0.51 \\
LLaMA-3 & ICL with sampling & 0.65 & 0.63 & 0.64 & 0.45 & \cellcolor{palegreen}0.54 & 0.49 \\

% Composite Fine-Tuning
Mistral & FT Composite & 0.81 & 0.82 & 0.82 & 0.35 & 0.45 & 0.39 \\
LLaMA-3 & FT Composite & \cellcolor{palegreen}0.87 & \cellcolor{palegreen}0.85 & \cellcolor{palegreen}0.86 & 0.35 & 0.33 & 0.34 \\

% Novel Fine-Tuning
Mistral & FT Novel & 0.46 & \cellcolor{palered}0.42 & \cellcolor{palered}0.44 & \cellcolor{palegreen}0.55 & \cellcolor{palegreen}0.54 & \cellcolor{palegreen}0.55 \\
LLaMA-3 & FT Novel & 0.47 & 0.43 & 0.45 & 0.54 & \cellcolor{palegreen}0.54 & 0.54 \\

% Blended Fine-Tuning
Mistral & FT Blended & 0.76 & 0.77 & 0.77 & 0.49 & 0.53 & 0.51 \\
LLaMA-3 & FT Blended & 0.77 & 0.74 & 0.76 & 0.52 & \cellcolor{palegreen}0.54 & 0.53 \\

% PyABSA
PyABSA & - & 0.77 & 0.75 & 0.76 & 0.33 & \cellcolor{palered}0.27 & 0.30 \\

\bottomrule
\end{tabular}
%\vspace{-0.5em}
\end{table}

We employed a paired bootstrap test, following the methodology of \citet{berg-kirkpatrick-etal-2012-empirical}, with \(10^5\) iterations to compute \(p\)-values. Results were deemed statistically significant for comparisons where \(p < 0.05\).

Open-weight LLMs' performance varies by dataset when used without fine-tuning. They performed worse than the PyABSA baseline on the Composite dataset (which matches PyABSA's training data), but outperformed it on the Novel dataset (which differs in domain and implicit aspect frequency). Using ICL with sampling significantly improved performance across both datasets, showing that providing relevant examples is an effective way to enhance LLMs' aspect extraction abilities.

Fine-tuning effectiveness depends on the similarity between training and evaluation data. When fine-tuned on the Composite dataset, both LLaMA-3 and Mistral showed significant performance gains on Composite samples compared to their non-fine-tuned versions, but their performance on Novel samples declined, falling below that of ICL with sampling. The reverse held true when fine-tuning on the Novel dataset: while significant improvements were observed on Novel samples, performance on Composite samples degraded below that of ICL with sampling. In contrast, fine-tuning on a mixed dataset combining both Novel and Composite samples yielded consistent performance gains across both dataset classes. Appendix~\ref{sec:expresults} presents detailed experimental results for individual datasets on the aspect extraction task, evaluated using both adjusted metrics \(\theta = 0.95\) and exact match criteria.

Table \ref{tab:detailed-results-asc} presents the experimental results for aspect sentiment classification (ASC) using standard precision (\(P\)), recall (\(R\)), and \(F_1\) metrics, as generalized metrics are unnecessary for this task.

\begin{table}[h]
\vspace{-5pt}
\centering
\scriptsize % Smaller font size
\setlength{\tabcolsep}{4pt} % Adjust column spacing
\renewcommand{\arraystretch}{1} % Less vertical space between rows
\caption{Experimental Results for Aspect Sentiment Classification.}
\label{tab:detailed-results-asc}

\begin{tabular}{@{}l@{\hspace{2pt}}c@{\hspace{5pt}}ccc@{\hspace{5pt}}ccc@{}}

\toprule
\multirow{2}{*}{\textbf{Model}} & \multirow{2}{*}{\textbf{Fine-Tuning / ICL}} & \multicolumn{3}{c}{\textbf{Composite Dataset}} & \multicolumn{3}{c}{\textbf{Novel Dataset}} \\
\cmidrule(lr){3-5} \cmidrule(lr){6-8}
& & \( P \) & \( R \) & \( F_1 \) & \( P \) & \( R \) & \( F_1 \) \\
\midrule
Mistral & Generic ICL & 0.55 & 0.33 & 0.41 & 0.56 & 0.28 & 0.37 \\
LLaMA-3 & Generic ICL & 0.53 & 0.33 & 0.38 & 0.59 & 0.36 & 0.43 \\

% Sampled ICL
Mistral & ICL with sampling & 0.56 & 0.36 & 0.44 & 0.58 & \cellcolor{palegreen}0.38 & 0.41 \\
LLaMA-3 & ICL with sampling & 0.55 & 0.35 & 0.43 & 0.59 & 0.36 & \cellcolor{palegreen}0.44 \\

% Composite Fine-Tuning
Mistral & FT Composite & 0.58 & 0.48 & 0.52 & 0.49 & 0.22 & 0.30 \\
LLaMA-3 & FT Composite & 0.60 & \cellcolor{palegreen}0.52 & \cellcolor{palegreen}0.56 & \cellcolor{palered}0.46 & 0.15 & 0.23 \\

% Novel Fine-Tuning
Mistral & FT Novel & 0.52 & \cellcolor{palered}0.24 & 0.31 & 0.60 & 0.33 & 0.43 \\
LLaMA-3 & FT Novel &\cellcolor{palered} 0.49 & \cellcolor{palered}0.24 & \cellcolor{palered}0.30 & \cellcolor{palegreen}0.69 & 0.31 & 0.42 \\

% Blended Fine-Tuning
Mistral & FT Blended & 0.58 & 0.46 & 0.51 & 0.65 & 0.31 & 0.42 \\
LLaMA-3 & FT Blended & 0.57 & 0.44 & 0.49 & 0.67 & 0.29 & 0.39 \\

% PyABSA
PyABSA & - & \cellcolor{palegreen}0.61 & 0.46 & 0.52 & 0.52 & \cellcolor{palered}0.14 & \cellcolor{palered}0.21 \\
\bottomrule
\end{tabular}
\vspace{-1pt}
\end{table}

ASC performance depends on successful aspect extraction, since only correctly identified aspects count toward recall and overall results. The patterns mirror aspect extraction findings: fine-tuning on a different dataset degrades model performance, whereas fine-tuning on similar data improves it. However, ICL with sampling showed no major improvement on the Novel dataset.

\section{Discussion and Further Research}
Our study reveals several key findings and corresponding future research directions. SOTA LLMs demonstrated effectiveness in generating initial annotations for the proposed dataset, despite inherent limitations like restricted context windows and occasional inaccuracies. The implemented multi-step annotation process, combining automated LLM-generated annotations with human validation, successfully streamlined the traditionally labor-intensive workflow while maintaining annotation quality through human oversight.

The employment of ICL with sampling proved effective for enhancing LLM performance in extracting ABSA pairs, offering advantages over fine-tuning approaches that can lead to overfitting and reduced generalizability. To build upon this success, future research should explore more sophisticated ICL strategies, such as retrieval-augmented ICL \cite{milios2023context}, which could further enhance the extraction of aspect-sentiment pairs.

Our proposed metric for generalized precision and recall captures model performance on the aspect extraction task while accounting for linguistic variability. Future work should focus on developing methods for automatic determination of the optimal threshold \(\theta\) value, investigating its relationship with various semantic similarity models. Additionally, implementing error detection methods could enable dynamic \(\theta\) adjustment, ensuring accurate performance measurement across both explicit and implicit aspect extraction scenarios.

Finally, adopting multi-step reasoning approaches like chain-of-thought prompting \cite{wei2022chain} or iterative refinement \cite{madaan2024self} presents a promising direction for improving both data annotation and pair extraction processes, potentially reducing the need for human intervention while maintaining output quality.

\section*{Conclusion}
This study serves as a proof of concept, demonstrating the applicability of our proposed approach in a challenging domain characterized by domain-specific terminology and a high prevalence of implicit aspects. While the dataset and findings are currently domain-specific, the methods introduced, such as the tailored evaluation metric and annotation framework, are designed to be adaptable to other contexts.

\section*{Limitations}

A significant drawback of employing LLMs for ABSA is the substantial computational resources required, particularly in terms of GPU usage. This demand can limit accessibility and scalability for practitioners with limited resources. However, as technological advancements continue to optimize hardware and algorithms, we anticipate a reduction in these computational barriers, potentially making LLM-based approaches the standard in ABSA.

The novel dataset is limited to a single domain and language (English), which may restrict its representativeness across other domains and languages. Additionally, it may not fully capture the richness and variability of natural language. Since it is generated by an LLM, it may exhibit limitations such as reduced lexical diversity and reliance on common phrasing patterns. Moreover, LLM-generated content may lack the contextual depth needed to capture implicit sentiment, aspect-specific variations, and the diversity of real-world expressions.

Annotation of datasets remains a considerable challenge. Identifying implicit aspects is a time-consuming and cognitively demanding task for human annotators. When aspects are abstract rather than concrete objects, inter-annotator agreement tends to decrease, affecting the reliability of the annotations. This highlights the need for improved annotation methodologies or assistance tools to better capture implicit aspects.

Moreover, we acknowledge that fine-tuning large language models on a small dataset, such as the 96 samples used in this study (25\% of the novel dataset), may not yield reliable or generalizable results. This limitation likely contributed to the observed decline in \(F_1\) scores on the composite dataset and the improvement on the novel dataset, suggesting potential overfitting. The large parameter space of LLMs necessitates substantial data for effective fine-tuning. To address this, future research should not only explore fine-tuning with larger, more diverse datasets, including those beyond peer-reviewed venues, but also incorporate regularization techniques such as dropout, weight decay, and early stopping.

Finally, the proposed metric for detecting aspects relies on the quality of the similarity scores, which may affect its consistency across different datasets.

\section*{Ethics Statement}

%\todo{ The authors also claim that they see no foreseeable risk of harm...this claim seems very strange since GPT-4 has over and over again been shown to generate biased data including sexist and racist output. So not only is the data possibly already biased, by using GPT-4 to evaluate itself, the data might become even more biased and introduce new types of bias into the system. The limitations section also does not in any way consider these issues. }
This research was conducted in accordance with the principles outlined in the ACL Code of Ethics, emphasizing honesty, transparency, and integrity throughout all stages of the study, from data collection to analysis and reporting. All data utilized in this study are publicly available and documented following best practices. 
%We also commit to publishing the code for the proposed metric upon the study's publication.

We acknowledge the potential biases introduced by using LLMs for data generation in creating this novel dataset. To ensure the quality of the synthetic dataset, an expert with a background in natural language processing conducted a comprehensive review of the generated content. This review focused on identifying potential biases introduced by the LLMs, including the over-representation of certain sentiment polarities, repetitive patterns in aspect phrasing, and cultural or linguistic biases. Based on the findings, approximately 6\% of the generated sentences were removed to address these concerns. Given the dataset's intended use in Aspect-Based Sentiment Analysis, we do not foresee significant risks of harm, affirming its utility for ethically advancing sentiment analysis research. Importantly, this dataset is not designed for tasks beyond ABSA.

Although large language models were employed in the initial data annotation step, subsequent steps were conducted by human annotators experienced in ABSA tasks and ethical considerations. This multi-step approach ensured a rigorous and ethically sound annotation process.


\bibliography{custom}

\begin{thebibliography}{60}
\expandafter\ifx\csname natexlab\endcsname\relax\def\natexlab#1{#1}\fi

\bibitem[{Bao et~al.(2023)Bao, Wang, Wang, Sun, Li, Mi, and Xu}]{bao-etal-2023-synthetic}
Jianzhu Bao, Rui Wang, Yasheng Wang, Aixin Sun, Yitong Li, Fei Mi, and Ruifeng Xu. 2023.
\newblock \href {https://doi.org/10.18653/v1/2023.acl-long.608} {A synthetic data generation framework for grounded dialogues}.
\newblock In \emph{Proceedings of the 61st Annual Meeting of the Association for Computational Linguistics (Volume 1: Long Papers)}, pages 10866--10882, Toronto, Canada. Association for Computational Linguistics.

\bibitem[{Berg-Kirkpatrick et~al.(2012)Berg-Kirkpatrick, Burkett, and Klein}]{berg-kirkpatrick-etal-2012-empirical}
Taylor Berg-Kirkpatrick, David Burkett, and Dan Klein. 2012.
\newblock \href {https://aclanthology.org/D12-1091} {An empirical investigation of statistical significance in {NLP}}.
\newblock In \emph{Proceedings of the 2012 Joint Conference on Empirical Methods in Natural Language Processing and Computational Natural Language Learning}, pages 995--1005, Jeju Island, Korea. Association for Computational Linguistics.

\bibitem[{Bhatt et~al.(2024)Bhatt, Chennabasappa, Li, Nikolaidis, Song, Wan, Ahmad, Aschermann, Chen, Kapil, Molnar, Whitman, and Saxe}]{bhatt2024cyberseceval}
Manish Bhatt, Sahana Chennabasappa, Yue Li, Cyrus Nikolaidis, Daniel Song, Shengye Wan, Faizan Ahmad, Cornelius Aschermann, Yaohui Chen, Dhaval Kapil, David Molnar, Spencer Whitman, and Joshua Saxe. 2024.
\newblock \href {http://arxiv.org/abs/2404.13161} {Cyberseceval 2: A wide-ranging cybersecurity evaluation suite for large language models}.

\bibitem[{Brauwers and Frasincar(2022)}]{10.1145/3503044}
Gianni Brauwers and Flavius Frasincar. 2022.
\newblock \href {https://doi.org/10.1145/3503044} {A survey on aspect-based sentiment classification}.
\newblock \emph{ACM Comput. Surv.}, 55(4).

\bibitem[{Brown et~al.(2020)Brown, Mann, Ryder, Subbiah, Kaplan, Dhariwal, Neelakantan, Shyam, Sastry, Askell, Agarwal, Herbert{-}Voss, Krueger, Henighan, Child, Ramesh, Ziegler, Wu, and et~al.}]{DBLP:journals/corr/abs-2005-14165}
Tom~B. Brown, Benjamin Mann, Nick Ryder, Melanie Subbiah, Jared Kaplan, Prafulla Dhariwal, Arvind Neelakantan, Pranav Shyam, Girish Sastry, Amanda Askell, Sandhini Agarwal, Ariel Herbert{-}Voss, Gretchen Krueger, Tom Henighan, Rewon Child, Aditya Ramesh, Daniel~M. Ziegler, Jeffrey Wu, and Clemens~Winter et~al. 2020.
\newblock \href {http://arxiv.org/abs/2005.14165} {Language models are few-shot learners}.
\newblock \emph{CoRR}, abs/2005.14165.

\bibitem[{Cai et~al.(2021)Cai, Xia, and Yu}]{cai2021aspect}
Hongjie Cai, Rui Xia, and Jianfei Yu. 2021.
\newblock Aspect-category-opinion-sentiment quadruple extraction with implicit aspects and opinions.
\newblock In \emph{Proceedings of the 59th Annual Meeting of the Association for Computational Linguistics and the 11th International Joint Conference on Natural Language Processing (Volume 1: Long Papers)}, pages 340--350.

\bibitem[{Chebolu et~al.(2023)Chebolu, Dernoncourt, Lipka, and Solorio}]{chebolu2023survey}
Siva Uday~Sampreeth Chebolu, Franck Dernoncourt, Nedim Lipka, and Thamar Solorio. 2023.
\newblock \href {http://arxiv.org/abs/2204.05232} {Survey of aspect-based sentiment analysis datasets}.

\bibitem[{Chebolu et~al.(2024)Chebolu, Dernoncourt, Lipka, and Solorio}]{oats}
Siva Uday~Sampreeth Chebolu, Franck Dernoncourt, Nedim Lipka, and Thamar Solorio. 2024.
\newblock \href {https://aclanthology.org/2024.lrec-main.1080/} {{OATS}: A challenge dataset for opinion aspect target sentiment joint detection for aspect-based sentiment analysis}.
\newblock In \emph{Proceedings of the 2024 Joint International Conference on Computational Linguistics, Language Resources and Evaluation (LREC-COLING 2024)}, pages 12336--12347, Torino, Italia. ELRA and ICCL.

\bibitem[{Crouse(2016)}]{Crouse}
David~F. Crouse. 2016.
\newblock \href {https://doi.org/10.1109/TAES.2016.140952} {On implementing 2d rectangular assignment algorithms}.
\newblock \emph{IEEE Transactions on Aerospace and Electronic Systems}, 52(4):1679--1696.

\bibitem[{Deng et~al.(2023)Deng, Bashlovkina, Han, Baumgartner, and Bendersky}]{deng2023llms}
Xiang Deng, Vasilisa Bashlovkina, Feng Han, Simon Baumgartner, and Michael Bendersky. 2023.
\newblock Llms to the moon? reddit market sentiment analysis with large language models.
\newblock In \emph{Companion Proceedings of the ACM Web Conference 2023}, pages 1014--1019.

\bibitem[{Dettmers et~al.(2024)Dettmers, Pagnoni, Holtzman, and Zettlemoyer}]{dettmers2024qlora}
Tim Dettmers, Artidoro Pagnoni, Ari Holtzman, and Luke Zettlemoyer. 2024.
\newblock Qlora: Efficient finetuning of quantized llms.
\newblock \emph{Advances in Neural Information Processing Systems}, 36.

\bibitem[{Dong et~al.(2014)Dong, Wei, Tan, Tang, Zhou, and Xu}]{dong2014adaptive}
Li~Dong, Furu Wei, Chuanqi Tan, Duyu Tang, Ming Zhou, and Ke~Xu. 2014.
\newblock Adaptive recursive neural network for target-dependent twitter sentiment classification.
\newblock In \emph{Proceedings of the 52nd annual meeting of the association for computational linguistics (volume 2: Short papers)}, pages 49--54.

\bibitem[{Dos~Santos et~al.(2021)Dos~Santos, Marcacini, and Rezende}]{dos2021multi}
Brucce~Neves Dos~Santos, Ricardo~Marcondes Marcacini, and Solange~Oliveira Rezende. 2021.
\newblock Multi-domain aspect extraction using bidirectional encoder representations from transformers.
\newblock \emph{IEEE Access}, 9:91604--91613.

\bibitem[{Eldan and Li(2023)}]{eldan2023tinystories}
Ronen Eldan and Yuanzhi Li. 2023.
\newblock \href {http://arxiv.org/abs/2305.07759} {Tinystories: How small can language models be and still speak coherent english?}

\bibitem[{Euzenat(2007)}]{euzenat2007semantic}
J{\'e}r{\^o}me Euzenat. 2007.
\newblock Semantic precision and recall for ontology alignment evaluation.
\newblock In \emph{Proc. 20th International Joint Conference on Artificial Intelligence (IJCAI)}, pages 348--353. AAAI Press.

\bibitem[{Gebru et~al.(2021)Gebru, Morgenstern, Vecchione, Vaughan, Wallach, Iii, and Crawford}]{gebru2021datasheets}
Timnit Gebru, Jamie Morgenstern, Briana Vecchione, Jennifer~Wortman Vaughan, Hanna Wallach, Hal~Daum{\'e} Iii, and Kate Crawford. 2021.
\newblock Datasheets for datasets.
\newblock \emph{Communications of the ACM}, 64(12):86--92.

\bibitem[{Goel et~al.(2023)Goel, Gueta, Gilon, Liu, Erell, Nguyen, Hao, Jaber, Reddy, Kartha, Steiner, Laish, and Feder}]{pmlr-v225-goel23a}
Akshay Goel, Almog Gueta, Omry Gilon, Chang Liu, Sofia Erell, Lan~Huong Nguyen, Xiaohong Hao, Bolous Jaber, Shashir Reddy, Rupesh Kartha, Jean Steiner, Itay Laish, and Amir Feder. 2023.
\newblock \href {https://proceedings.mlr.press/v225/goel23a.html} {Llms accelerate annotation for medical information extraction}.
\newblock In \emph{Proceedings of the 3rd Machine Learning for Health Symposium}, volume 225 of \emph{Proceedings of Machine Learning Research}, pages 82--100. PMLR.

\bibitem[{Gray et~al.(2023)Gray, Savelka, Oliver, and Ashley}]{gray2023can}
Morgan Gray, Jaromir Savelka, Wesley Oliver, and Kevin Ashley. 2023.
\newblock Can gpt alleviate the burden of annotation?
\newblock In \emph{Legal Knowledge and Information Systems}, pages 157--166. IOS Press.

\bibitem[{He et~al.(2024)He, Huang, Ding, Rohatgi, and Huang}]{gpt4annot}
Zeyu He, Chieh-Yang Huang, Chien-Kuang~Cornelia Ding, Shaurya Rohatgi, and Ting-Hao~Kenneth Huang. 2024.
\newblock \href {https://doi.org/10.1145/3613904.3642834} {If in a crowdsourced data annotation pipeline, a gpt-4}.
\newblock In \emph{Proceedings of the 2024 CHI Conference on Human Factors in Computing Systems}, CHI '24, New York, NY, USA. Association for Computing Machinery.

\bibitem[{Hu et~al.(2021)Hu, Shen, Wallis, Allen-Zhu, Li, Wang, Wang, and Chen}]{hu2021lora}
Edward~J. Hu, Yelong Shen, Phillip Wallis, Zeyuan Allen-Zhu, Yuanzhi Li, Shean Wang, Lu~Wang, and Weizhu Chen. 2021.
\newblock \href {http://arxiv.org/abs/2106.09685} {Lora: Low-rank adaptation of large language models}.

\bibitem[{Jiang et~al.(2023)Jiang, Sablayrolles, Mensch, Bamford, Chaplot, de~las Casas, Bressand, Lengyel, Lample, Saulnier, Lavaud, Lachaux, Stock, Scao, Lavril, Wang, Lacroix, and Sayed}]{jiang2023mistral}
Albert~Q. Jiang, Alexandre Sablayrolles, Arthur Mensch, Chris Bamford, Devendra~Singh Chaplot, Diego de~las Casas, Florian Bressand, Gianna Lengyel, Guillaume Lample, Lucile Saulnier, Lélio~Renard Lavaud, Marie-Anne Lachaux, Pierre Stock, Teven~Le Scao, Thibaut Lavril, Thomas Wang, Timothée Lacroix, and William~El Sayed. 2023.
\newblock \href {http://arxiv.org/abs/2310.06825} {Mistral 7b}.

\bibitem[{Jiang et~al.(2019)Jiang, Chen, Xu, Ao, and Yang}]{jiang-etal-2019-challenge}
Qingnan Jiang, Lei Chen, Ruifeng Xu, Xiang Ao, and Min Yang. 2019.
\newblock \href {https://doi.org/10.18653/v1/D19-1654} {A challenge dataset and effective models for aspect-based sentiment analysis}.
\newblock In \emph{Proceedings of the 2019 Conference on Empirical Methods in Natural Language Processing and the 9th International Joint Conference on Natural Language Processing (EMNLP-IJCNLP)}, pages 6280--6285, Hong Kong, China. Association for Computational Linguistics.

\bibitem[{Kheiri and Karimi(2023)}]{kheiri2023sentimentgpt}
Kiana Kheiri and Hamid Karimi. 2023.
\newblock Sentimentgpt: Exploiting gpt for advanced sentiment analysis and its departure from current machine learning.
\newblock \emph{arXiv preprint arXiv:2307.10234}.

\bibitem[{Kramchaninova and Defauw(2022)}]{kramchaninova2022synthetic}
Alina Kramchaninova and Arne Defauw. 2022.
\newblock Synthetic data generation for multilingual domain-adaptable question answering systems.
\newblock In \emph{Proceedings of the 23rd Annual Conference of the European Association for Machine Translation}, pages 151--160.

\bibitem[{Krugmann and Hartmann(2024)}]{krugmann2024sentiment}
Jan~Ole Krugmann and Jochen Hartmann. 2024.
\newblock Sentiment analysis in the age of generative ai.
\newblock \emph{Customer Needs and Solutions}, 11(1):3.

\bibitem[{Liu(2012)}]{liu2022sentiment}
Bing Liu. 2012.
\newblock \href {https://doi.org/10.1007/978-3-031-02145-9} {\emph{Sentiment Analysis and Opinion Mining}}.
\newblock Synthesis Lectures on Human Language Technologies. Springer.

\bibitem[{Liyanage et~al.(2024)Liyanage, Gokani, and Mago}]{liyanage2024gpt}
Chandreen~R Liyanage, Ravi Gokani, and Vijay Mago. 2024.
\newblock Gpt-4 as an x data annotator: Unraveling its performance on a stance classification task.
\newblock \emph{PloS one}, 19(8):e0307741.

\bibitem[{Ma et~al.(2019)Ma, Li, Wu, Xie, and Wang}]{ma2019exploring}
Dehong Ma, Sujian Li, Fangzhao Wu, Xing Xie, and Houfeng Wang. 2019.
\newblock Exploring sequence-to-sequence learning in aspect term extraction.
\newblock In \emph{Proceedings of the 57th annual meeting of the association for computational linguistics}, pages 3538--3547.

\bibitem[{Ma et~al.(2023)Ma, Hu, Liu, Yang, Li, Yu, and Wen}]{ma-etal-2023-amr}
Fukun Ma, Xuming Hu, Aiwei Liu, Yawen Yang, Shuang Li, Philip~S. Yu, and Lijie Wen. 2023.
\newblock \href {https://doi.org/10.18653/v1/2023.acl-long.19} {{AMR}-based network for aspect-based sentiment analysis}.
\newblock In \emph{Proceedings of the 61st Annual Meeting of the Association for Computational Linguistics (Volume 1: Long Papers)}, pages 322--337, Toronto, Canada. Association for Computational Linguistics.

\bibitem[{Madaan et~al.(2024)Madaan, Tandon, Gupta, Hallinan, Gao, Wiegreffe, Alon, Dziri, Prabhumoye, Yang et~al.}]{madaan2024self}
Aman Madaan, Niket Tandon, Prakhar Gupta, Skyler Hallinan, Luyu Gao, Sarah Wiegreffe, Uri Alon, Nouha Dziri, Shrimai Prabhumoye, Yiming Yang, et~al. 2024.
\newblock Self-refine: Iterative refinement with self-feedback.
\newblock \emph{Advances in Neural Information Processing Systems}, 36.

\bibitem[{Meng et~al.(2019)Meng, Wei, Liu, Zhu, and Yin}]{MengABSA}
Wei Meng, Yongqing Wei, Peiyu Liu, Zhenfang Zhu, and Hongxia Yin. 2019.
\newblock \href {https://doi.org/10.1109/ACCESS.2019.2952888} {Aspect based sentiment analysis with feature enhanced attention cnn-bilstm}.
\newblock \emph{IEEE Access}, 7:167240--167249.

\bibitem[{Meyer et~al.(2022)Meyer, Elsweiler, Ludwig, Fernandez-Pichel, and Losada}]{meyer2022we}
Selina Meyer, David Elsweiler, Bernd Ludwig, Marcos Fernandez-Pichel, and David~E Losada. 2022.
\newblock Do we still need human assessors? prompt-based gpt-3 user simulation in conversational ai.
\newblock In \emph{Proceedings of the 4th Conference on Conversational User Interfaces}, pages 1--6.

\bibitem[{Meškelė and Frasincar(2020)}]{MESKELE2020102211}
Donatas Meškelė and Flavius Frasincar. 2020.
\newblock \href {https://doi.org/https://doi.org/10.1016/j.ipm.2020.102211} {Aldonar: A hybrid solution for sentence-level aspect-based sentiment analysis using a lexicalized domain ontology and a regularized neural attention model}.
\newblock \emph{Information Processing and Management}, 57(3):102211.

\bibitem[{Milios et~al.(2023)Milios, Reddy, and Bahdanau}]{milios2023context}
Aristides Milios, Siva Reddy, and Dzmitry Bahdanau. 2023.
\newblock In-context learning for text classification with many labels.
\newblock In \emph{GenBench: The first workshop on generalisation (benchmarking) in NLP}, page 173.

\bibitem[{Mirowski et~al.(2023)Mirowski, Mathewson, Pittman, and Evans}]{Mirowski}
Piotr Mirowski, Kory~W. Mathewson, Jaylen Pittman, and Richard Evans. 2023.
\newblock \href {https://doi.org/10.1145/3544548.3581225} {Co-writing screenplays and theatre scripts with language models: Evaluation by industry professionals}.
\newblock In \emph{Proceedings of the 2023 CHI Conference on Human Factors in Computing Systems}, CHI '23, New York, NY, USA. Association for Computing Machinery.

\bibitem[{Mohta et~al.(2023)Mohta, Ak, Xu, and Shen}]{pmlr-v239-mohta23a}
Jay Mohta, Kenan Ak, Yan Xu, and Mingwei Shen. 2023.
\newblock \href {https://proceedings.mlr.press/v239/mohta23a.html} {Are large language models good annotators?}
\newblock In \emph{Proceedings on "I Can't Believe It's Not Better: Failure Modes in the Age of Foundation Models" at NeurIPS 2023 Workshops}, volume 239 of \emph{Proceedings of Machine Learning Research}, pages 38--48. PMLR.

\bibitem[{Muennighoff et~al.(2023)Muennighoff, Tazi, Magne, and Reimers}]{mteb}
Niklas Muennighoff, Nouamane Tazi, Loic Magne, and Nils Reimers. 2023.
\newblock \href {https://doi.org/10.18653/v1/2023.eacl-main.148} {{MTEB}: Massive text embedding benchmark}.
\newblock In \emph{Proceedings of the 17th Conference of the European Chapter of the Association for Computational Linguistics}, pages 2014--2037, Dubrovnik, Croatia. Association for Computational Linguistics.

\bibitem[{Nazir et~al.(2020)Nazir, Rao, Wu, and Sun}]{nazir2020issues}
Ambreen Nazir, Yuan Rao, Lianwei Wu, and Ling Sun. 2020.
\newblock Issues and challenges of aspect-based sentiment analysis: A comprehensive survey.
\newblock \emph{IEEE Transactions on Affective Computing}, 13(2):845--863.

\bibitem[{Ni et~al.(2021)Ni, Ábrego, Constant, Ma, Hall, Cer, and Yang}]{ni2021sentencet5}
Jianmo Ni, Gustavo~Hernández Ábrego, Noah Constant, Ji~Ma, Keith~B. Hall, Daniel Cer, and Yinfei Yang. 2021.
\newblock \href {http://arxiv.org/abs/2108.08877} {Sentence-t5: Scalable sentence encoders from pre-trained text-to-text models}.

\bibitem[{OpenAI(2023)}]{openai2023gpt4}
OpenAI. 2023.
\newblock \href {http://arxiv.org/abs/2303.08774} {Gpt-4 technical report}.

\bibitem[{Pontiki et~al.(2015)Pontiki, Galanis, Papageorgiou, Manandhar, and Androutsopoulos}]{pontiki-etal-2015-semeval}
Maria Pontiki, Dimitris Galanis, Haris Papageorgiou, Suresh Manandhar, and Ion Androutsopoulos. 2015.
\newblock \href {https://doi.org/10.18653/v1/S15-2082} {{S}em{E}val-2015 task 12: Aspect based sentiment analysis}.
\newblock In \emph{Proceedings of the 9th International Workshop on Semantic Evaluation ({S}em{E}val 2015)}, pages 486--495, Denver, Colorado. Association for Computational Linguistics.

\bibitem[{Pontiki et~al.(2014)Pontiki, Galanis, Pavlopoulos, Papageorgiou, Androutsopoulos, and Manandhar}]{pontiki-etal-2014-semeval}
Maria Pontiki, Dimitris Galanis, John Pavlopoulos, Harris Papageorgiou, Ion Androutsopoulos, and Suresh Manandhar. 2014.
\newblock \href {https://doi.org/10.3115/v1/S14-2004} {{S}em{E}val-2014 task 4: Aspect based sentiment analysis}.
\newblock In \emph{Proceedings of the 8th International Workshop on Semantic Evaluation ({S}em{E}val 2014)}, pages 27--35, Dublin, Ireland. Association for Computational Linguistics.

\bibitem[{Scaria et~al.(2024)Scaria, Gupta, Goyal, Sawant, Mishra, and Baral}]{scaria2023instructabsa}
Kevin Scaria, Himanshu Gupta, Siddharth Goyal, Saurabh Sawant, Swaroop Mishra, and Chitta Baral. 2024.
\newblock \href {https://doi.org/10.18653/v1/2024.naacl-short.63} {{I}nstruct{ABSA}: Instruction learning for aspect based sentiment analysis}.
\newblock In \emph{Proceedings of the 2024 Conference of the North American Chapter of the Association for Computational Linguistics: Human Language Technologies (Volume 2: Short Papers)}, pages 720--736, Mexico City, Mexico. Association for Computational Linguistics.

\bibitem[{{\v{S}}m{\'\i}d et~al.(2024){\v{S}}m{\'\i}d, Priban, and Kral}]{smid-etal-2024-llama}
Jakub {\v{S}}m{\'\i}d, Pavel Priban, and Pavel Kral. 2024.
\newblock \href {https://aclanthology.org/2024.wassa-1.6} {{LL}a{MA}-based models for aspect-based sentiment analysis}.
\newblock In \emph{Proceedings of the 14th Workshop on Computational Approaches to Subjectivity, Sentiment, {\&} Social Media Analysis}, pages 63--70, Bangkok, Thailand. Association for Computational Linguistics.

\bibitem[{Tian et~al.(2024)Tian, Liu, Song, Xia, and Zhang}]{tian-etal-2024-aspect}
Yuanhe Tian, Chang Liu, Yan Song, Fei Xia, and Yongdong Zhang. 2024.
\newblock \href {https://doi.org/10.18653/v1/2024.findings-naacl.194} {Aspect-based sentiment analysis with context denoising}.
\newblock In \emph{Findings of the Association for Computational Linguistics: NAACL 2024}, pages 3083--3095, Mexico City, Mexico. Association for Computational Linguistics.

\bibitem[{Tubishat et~al.(2018)Tubishat, Idris, and Abushariah}]{TUBISHAT2018545}
Mohammad Tubishat, Norisma Idris, and Mohammad~A.M. Abushariah. 2018.
\newblock \href {https://doi.org/https://doi.org/10.1016/j.ipm.2018.03.008} {Implicit aspect extraction in sentiment analysis: Review, taxonomy, oppportunities, and open challenges}.
\newblock \emph{Information Processing and Management}, 54(4):545--563.

\bibitem[{Wang et~al.(2024{\natexlab{a}})Wang, He, Li, Chen, Li, Han, Teng, and Ji}]{wang2024refining}
Haining Wang, Kang He, Bobo Li, Lei Chen, Fei Li, Xu~Han, Chong Teng, and Donghong Ji. 2024{\natexlab{a}}.
\newblock Refining and synthesis: A simple yet effective data augmentation framework for cross-domain aspect-based sentiment analysis.
\newblock In \emph{Findings of the Association for Computational Linguistics ACL 2024}, pages 10318--10329.

\bibitem[{Wang et~al.(2020)Wang, Shen, Yang, Quan, and Wang}]{wang2020relational}
Kai Wang, Weizhou Shen, Yunyi Yang, Xiaojun Quan, and Rui Wang. 2020.
\newblock \href {http://arxiv.org/abs/2004.12362} {Relational graph attention network for aspect-based sentiment analysis}.

\bibitem[{Wang et~al.(2024{\natexlab{b}})Wang, Zhang, Yang, Guo, and Li}]{wang-etal-2024-dagcn}
Zhihao Wang, Bo~Zhang, Ru~Yang, Chang Guo, and Maozhen Li. 2024{\natexlab{b}}.
\newblock \href {https://doi.org/10.18653/v1/2024.findings-naacl.120} {{DAGCN}: Distance-based and aspect-oriented graph convolutional network for aspect-based sentiment analysis}.
\newblock In \emph{Findings of the Association for Computational Linguistics: NAACL 2024}, pages 1863--1876, Mexico City, Mexico. Association for Computational Linguistics.

\bibitem[{Wankhade et~al.(2022)Wankhade, Rao, and Kulkarni}]{wankhade2022survey}
Mayur Wankhade, Annavarapu Chandra~Sekhara Rao, and Chaitanya Kulkarni. 2022.
\newblock A survey on sentiment analysis methods, applications, and challenges.
\newblock \emph{Artificial Intelligence Review}, 55(7):5731--5780.

\bibitem[{Wei et~al.(2022)Wei, Wang, Schuurmans, Bosma, Xia, Chi, Le, Zhou et~al.}]{wei2022chain}
Jason Wei, Xuezhi Wang, Dale Schuurmans, Maarten Bosma, Fei Xia, Ed~Chi, Quoc~V Le, Denny Zhou, et~al. 2022.
\newblock Chain-of-thought prompting elicits reasoning in large language models.
\newblock \emph{Advances in neural information processing systems}, 35:24824--24837.

\bibitem[{Wu et~al.(2022)Wu, Zhang, Shi, Wu, and Song}]{WU2022107736}
Haiyan Wu, Zhiqiang Zhang, Shaoyun Shi, Qingfeng Wu, and Haiyu Song. 2022.
\newblock \href {https://doi.org/https://doi.org/10.1016/j.knosys.2021.107736} {Phrase dependency relational graph attention network for aspect-based sentiment analysis}.
\newblock \emph{Knowledge-Based Systems}, 236:107736.

\bibitem[{Xu et~al.(2020)Xu, Wang, Yang, and Kang}]{xu2020target}
Borun Xu, Xiaoxiao Wang, Bo~Yang, and Zhongfeng Kang. 2020.
\newblock Target embedding and position attention with lstm for aspect based sentiment analysis.
\newblock In \emph{Proceedings of the 2020 5th International Conference on Mathematics and Artificial Intelligence}, pages 93--97.

\bibitem[{Yang et~al.(2023)Yang, Zhang, and Li}]{pyabsa}
Heng Yang, Chen Zhang, and Ke~Li. 2023.
\newblock \href {https://doi.org/10.1145/3583780.3614752} {Pyabsa: A modularized framework for reproducible aspect-based sentiment analysis}.
\newblock In \emph{Proceedings of the 32nd ACM International Conference on Information and Knowledge Management}, CIKM '23, page 5117–5122, New York, NY, USA. Association for Computing Machinery.

\bibitem[{Yang et~al.(2024)Yang, Jiang, Zhao, Zeng, Liu, and Jia}]{yang2024faima}
Songhua Yang, Xinke Jiang, Hanjie Zhao, Wenxuan Zeng, Hongde Liu, and Yuxiang Jia. 2024.
\newblock \href {http://arxiv.org/abs/2403.01063} {Faima: Feature-aware in-context learning for multi-domain aspect-based sentiment analysis}.

\bibitem[{Yu et~al.(2023)Yu, Zhao, and Xia}]{yu2023cross}
Jianfei Yu, Qiankun Zhao, and Rui Xia. 2023.
\newblock Cross-domain data augmentation with domain-adaptive language modeling for aspect-based sentiment analysis.
\newblock In \emph{Proceedings of the 61st Annual Meeting of the Association for Computational Linguistics (Volume 1: Long Papers)}, pages 1456--1470.

\bibitem[{Zhang et~al.(2023{\natexlab{a}})Zhang, Zhu, Liu, Bao, Wu, Sun, and Xu}]{zhang-etal-2023-span}
Mao Zhang, Yongxin Zhu, Zhen Liu, Zhimin Bao, Yunfei Wu, Xing Sun, and Linli Xu. 2023{\natexlab{a}}.
\newblock \href {https://doi.org/10.18653/v1/2023.acl-long.515} {Span-level aspect-based sentiment analysis via table filling}.
\newblock In \emph{Proceedings of the 61st Annual Meeting of the Association for Computational Linguistics (Volume 1: Long Papers)}, pages 9273--9284, Toronto, Canada. Association for Computational Linguistics.

\bibitem[{Zhang et~al.(2023{\natexlab{b}})Zhang, Li, Deng, Bing, and Lam}]{WOS:001089176900009}
Wenxuan Zhang, Xin Li, Yang Deng, Lidong Bing, and Wai Lam. 2023{\natexlab{b}}.
\newblock \href {https://doi.org/10.1109/TKDE.2022.3230975} {A survey on aspect-based sentiment analysis: Tasks, methods, and challenges}.
\newblock \emph{IEEE TRANSACTIONS ON KNOWLEDGE AND DATA ENGINEERING}, 35(11):11019--11038.

\bibitem[{Zhou et~al.(2020)Zhou, Huang, Hu, and He}]{huang}
Jie Zhou, Jimmy~Xiangji Huang, Qinmin~Vivian Hu, and Liang He. 2020.
\newblock \href {https://doi.org/https://doi.org/10.1016/j.knosys.2020.106292} {Sk-gcn: Modeling syntax and knowledge via graph convolutional network for aspect-level sentiment classification}.
\newblock \emph{Knowledge-Based Systems}, 205:106292.

\bibitem[{Zhou et~al.(2024)Zhou, Shen, and Li}]{zhou-etal-2024-gcnet}
Ting Zhou, Ying Shen, and Yinghui Li. 2024.
\newblock \href {https://aclanthology.org/2024.lrec-main.669} {{GCN}et: Global-and-context collaborative learning for aspect-based sentiment analysis}.
\newblock In \emph{Proceedings of the 2024 Joint International Conference on Computational Linguistics, Language Resources and Evaluation (LREC-COLING 2024)}, pages 7570--7580, Torino, Italia. ELRA and ICCL.

\end{thebibliography}
\bibliographystyle{acl_natbib}

\appendix

\onecolumn

\section{Examples of Prompts and Generated Feedback}
\label{sec:promts}
\scriptsize
\begin{tcolorbox}[colback=gray!5, colframe=gray!80, title=Prompt 1]
You are a human who participated as a volunteer in New Brunswick sport games for minorities (the event is called NBMSG). You are invited to take a survey. Generate 10 naturally sounding diverse responses with variable length (from 5 to 10 sentences with average of 6-7 sentences, try to provide longer responses). Try to be informal and use a lot of slang.
\end{tcolorbox}

\begin{tcolorbox}[colback=white, colframe=black, title=Chunk Of The Model Response (Gemini)]
1.  Sure ya! I'd be happy to answer some questions about NBMSG. Those games were AWESOME!  Met so many cool ppl and really felt like the volunteers were appreciated.  Although, maybe next time a bit more info on what to expect beforehand woulda been helpful. 

2.  Hey yeah, I can totally do a survey!  NBMSG was a blastt  - so glad I volunteered.  It was tiring at times, but seeing those kids faces light up made it all worth it.  Defo considering doing it again next year!

3.  Absolutely, hit me with that survey!  Being a volunteer at NBMSG was an amazing experience.  It was so inspiring to see how passionate everyone was about sports and inclusion.  Just a suggestion, maybe a little more organization with scheduling would be good? But overall, 10/10 would volunteer again!  

4.  Ya ofc I'll do the survey!  Those NBMSG games were lit.  I volunteered at the concession stand and it was craaazy busy at times, but the teamwork between us volunteers was awesome .  Definitely recommend volunteering there if you get the chance!

\end{tcolorbox}

\begin{tcolorbox}[colback=gray!5, colframe=gray!80, title=Prompt 2]
You are a human who participated as a volunteer in New Brunswick sport games for minorities (the event is called NBMSG). You are invited to take a survey. Generate 20 naturally sounding diverse responses with variable length (from 1 to 5 sentences with average of 3-5 sentences, try to provide longer responses). Use a lot of typos, grammatical errors, slang, emojis, etc.. Be negative.
\end{tcolorbox}

\begin{tcolorbox}[colback=white, colframe=black, title=Chunk Of The Model Response (Gemini - Emojis Removed)]
1. Ugh, finally got around to this NBMSG survey.  Took forever to find the time between work and everything, and let's just say it wasn't the best volunteer experience.  Not sure what I expected, but maybe a bit more organization next time?

2.  Fine, here's my survey response about NBMSG.  Honestly, kind of disappointed.  The description made it sound way more fun and impactful than it actually was.  Definitely not what I signed up for.

3.  **NBMSG survey**...  Not sure what I was thinking volunteering there.  Felt kinda lost and confused most of the time.  Barely any training, and nobody seemed to know what was going on half the time.  Doubt I'll be back next year. 

4.  Ugh, this NBMSG survey is taking foreverrrrrrrr.   Get on with it already!  Look, I volunteered my time, the least they could do is make this quick and painless.

\end{tcolorbox}

\begin{tcolorbox}[colback=gray!5, colframe=gray!80, title=Prompt 3]
You are a human who participated as a volunteer in New Brunswick sport games for minorities (the event is called NBMSG). You are invited to take a survey. Generate 20 naturally sounding diverse responses with variable length (from 1 to 10 sentences with average of 4-6 sentences, try to provide longer responses). Focus on negative sentiment towards various aspects. Be emotional and creative. Don't hesitate to use slang.
\end{tcolorbox}

\begin{tcolorbox}[colback=white, colframe=black, title=Chunk Of The Model Response (GPT-4)]
1. "The IT setup was a hot mess – like, seriously? In the age of tech, how are we dealing with systems crashing left and right? Made it impossible to keep up with schedules and scores. Frustrating doesn't even start to cover it."

2. "Food options were straight-up disappointing. With such a rich tapestry of cultures, I expected a feast for the senses. Instead, we got the same old, same old. Missed opportunity to showcase the diversity we were all there to celebrate."

3. "Weather threw us for a loop, and it felt like the organizers hadn't even glanced at the forecast. Events got cancelled and shuffled around with hardly any heads-up. Chaos isn't even the word for it."

4. "Volunteer coordination was a joke. Half the time, I didn't know where I was supposed to be or what I was supposed to be doing. It was like trying to solve a puzzle without the picture. Super frustrating."

\end{tcolorbox}

\begin{tcolorbox}[colback=gray!5, colframe=gray!80, title=Prompt 4]
You are a human who participated as a volunteer in New Brunswick sport games for minorities (the event is called NBMSG). 
You are invited to take a survey. Generate 20 naturally sounding diverse responses with variable length (from 1 to 10 sentences with average of 4-6 sentences, try to provide longer responses). Focus on positive and neutral sentiment towards various aspects. 
The style of comments should reflect the following person: 35-50 years old, education: university degree
\end{tcolorbox}

\begin{tcolorbox}[colback=white, colframe=black, title=Chunk Of The Model Response (GPT-4)]

1. "Volunteering at NBMSG provided a wonderful opportunity to meet people from various backgrounds and professions. The diversity among the volunteers alone was a testament to the inclusive nature of the event. It was an enriching experience that I believe contributed positively to the community spirit."

2. "Participating as a volunteer at NBMSG was an eye-opening experience. It was refreshing to see such a wide range of ages, backgrounds, and abilities come together in celebration of sport and culture. The positive atmosphere was contagious and made the hard work worthwhile."

3. "The event's use of digital platforms for scheduling and updates was a forward-thinking approach, though not without its challenges. As we become increasingly reliant on technology, it's essential for events like NBMSG to continue improving their digital infrastructure to enhance the experience for all involved."

4. "The logistical coordination of NBMSG, while complex, was handled with notable effort and dedication by the organizers. As a volunteer, I felt supported and appreciated, which in turn motivated me to contribute my best to the event's success. It was a valuable learning experience."

\end{tcolorbox}
\newpage

\newpage
\twocolumn
\section{Dataset Annotation Process}
\label{sec:annotproc}
\normalsize

In the first step of dataset annotation, both GPT-4 and Gemini 1.0 Ultra were tasked with annotating the data. Appendix~\ref{sec:annotations} provides sample annotation prompts and responses from the models.

\begin{figure}[htbp] % Adjust placement options as needed
    \centering 
    %\hspace*{-13pt} % Adjust the value as needed to move the figure left
    \includegraphics[width=0.5\textwidth]{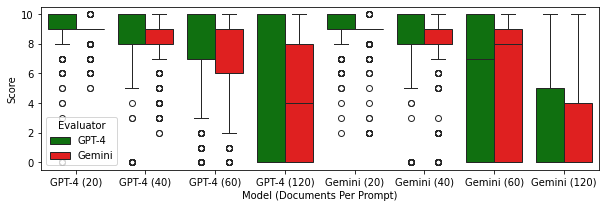}
    \caption{Impact of Document Quantity on Annotation Quality: Evaluation Scores from GPT-4 and Gemini Models.}
    \label{fig:ctx_vs_nchar} 
\end{figure}

As an auxiliary step (indicated by the blue box in Figure~\ref{fig:annotation} of the main text), we experimented with varying the number of documents per prompt, ranging from 20 to 120, to assess how this variation affects annotation quality. The results indicated that the quality of annotations for both models substantially decreased as the number of documents per prompt increased. To quantify this, we asked both models to evaluate the annotation sets produced with 20, 40, 60, and 120 documents per prompt using a scale from 1 to 10. This scale was chosen to provide a sufficiently granular assessment while maintaining simplicity for quantitative interpretation. Notably, the models were unaware of both the number of documents per prompt and which model had provided the annotations. Figure~\ref{fig:ctx_vs_nchar} illustrates the evaluation scores for different numbers of documents per prompt (20, 40, 60, 120). The y-axis shows the score distribution for GPT-4 (green boxplots) and Gemini (red boxplots), while the x-axis represents the annotations provided by the models, with the corresponding number of documents per prompt indicated in parentheses. The mean Fleiss' Kappa, calculated across four binned labels, is 0.62.

%\vspace{-3pt}
\captionsetup{skip=0pt} 

The second step involved refining the annotations. Two annotation sets were selected for this purpose: one from GPT-4 and another from Gemini, each generated with 20 documents per prompt. Three undergraduate student volunteers, familiar with ABSA tasks, were tasked with selecting the most suitable annotation from each set based on the accuracy of identified aspects and their polarities. This evaluation yielded a Fleiss' Kappa score of 0.3, reflecting the inherent difficulty of implicit aspect identification and the subjective nature of interpreting subtle or context-dependent aspects. This highlights the importance of the third step involving thorough expert review to ensure the quality of the final annotations. Appendix~\ref{sec:volinstr} provides the written instructions given to volunteers, along with details of the training sessions provided.

%The author mentions selecting appropriate annotations through volunteers and experts, but there is no introduction of the experts' professional qualifications, nor is it clear how the expert-filtered data is handled.
The third step involved revision and adjustment by an expert\footnote{Holds MSc in Computer Science}, who selected the annotations based on the volunteers' feedback and their own judgment, particularly in cases with low volunteer agreement. Adjustments were required for 12.5\% of the documents.

\newpage
\onecolumn
\section{Examples of Annotation Prompts and Generated Annotations}
\label{sec:annotations}
\tiny
\begin{tcolorbox}[colback=gray!5, colframe=gray!80, title=Annotation Prompt 1]
Follow the instructions precisely. 
Provide answers as directed in the example below (key-value pairs in curly braces, separated by comma, do not reprint sentences and do not provide any additional information). 
Do not divide answers into categories, just follow the sequence of sentences.

Given the following feedback from volunteers of an event called NBMSG, perform aspect-based sentiment analysis: identify aspects and polarities (Positive, Negative, Neutral) as in the examples below. 

Note: the empty dictionary for the third example indicates that there are no aspects or polarities associated with the text:

1 "I like school but the organization of the art classes needs improvement."

1 \{"school":"Positive","organization of art classes":"Negative"\}

2 "Dog drinks water."

2 \{\}

3 "Volunteers could be allocated better, but at least everyone seemed to be fine with that."

3 \{"Allocation of volunteers":"Neutral"\}

Feedback:

1. "The organization was a total mess, honestly. They had us running around with barely any direction. And when you ask for help, it's like no one knew what was going on either."

2. "I was really excited to help out, but the weather totally ruined it for me. It was like they didn't even check the forecast. We were all soaked and miserable."

3. "The sign-up process was a nightmare. The website kept crashing, and when it didn't, it was super slow. They really need to get their IT stuff together."

4. "Food there was just sad. Like, if you're gonna have us there all day, at least provide something decent to eat. The options were super limited and way overpriced."

5. "Nobody mentioned anything about the parking situation. I ended up parking a mile away and walking. They should've sent out some info or something."

6. "They said there'd be Wi-Fi, but I couldn't connect the whole time I was there. Made it impossible to post anything or stay updated."

7. "I felt really unprepared. They didn't give us enough info before the event, so I was kinda just winging it the whole time. Not cool."

8. "It's like they didn't even think about the weather. We were out in the sun with no shade for hours. I got a wicked sunburn."

9. "The equipment was pretty outdated. It's supposed to be a sports event, right? Shouldn't they have decent gear for the athletes?"

10. "I was hyped to help out, but the lack of organization just killed the vibe. Waited around a lot for someone to tell me what to do."

11. "The schedule was a joke. Everything was running late or overlapping. Made it impossible to catch the events I wanted to see."

12. "There was literally nowhere to get water unless you wanted to buy overpriced bottles. For an outdoor event, that's just irresponsible."

13. "Communication was a major issue. Half the time, the announcements were so muffled you couldn't understand anything."

14. "They promised some cool activities, but a bunch got canceled last minute. Super disappointing for everyone looking forward to them."

15. "I wanted to contribute, but the disorganization made it so stressful. At one point, I just sat down because I had no clue where I was needed."

16. "The volunteer shirts were ugly and uncomfortable. It's a small thing, but if I'm wearing it all day, at least make it bearable."

17. "The medical tent was hard to find, and when you did, they seemed really unprepared for any serious issues. That's worrying at a sports event."

18. "They were super strict about bringing your own food, which wouldn't have been a problem if their food wasn't so bad. I was starving by the end."

19. "I felt like there was a lack of appreciation for the volunteers. A simple 'thank you' would have gone a long way."

20. "The event was supposed to be about inclusivity, but it felt like they missed the mark on several fronts. From accessibility issues to lack of diversity in food choices."
\end{tcolorbox}

\begin{tcolorbox}[colback=white, colframe=black, title=Model Response (GPT-4)]
1. \{"Organization":"Negative"\}

2. \{"Weather preparedness":"Negative"\}

3. \{"Sign-up process":"Negative", "Website functionality":"Negative"\}

4. \{"Food quality":"Negative", "Food pricing":"Negative"\}

5. \{"Parking information":"Negative"\}

6. \{"Wi-Fi availability":"Negative"\}

7. \{"Pre-event information":"Negative"\}

8. \{"Weather preparedness":"Negative", "Sun protection":"Negative"\}

9. \{"Equipment quality":"Negative"\}

10. \{"Organization":"Negative"\}

11. \{"Schedule":"Negative"\}

12. \{"Water availability":"Negative"\}

13. \{"Communication clarity":"Negative"\}

14. \{"Activity fulfillment":"Negative"\}

15. \{"Organization":"Negative"\}

16. \{"Volunteer shirts":"Negative"\}

17. \{"Medical preparedness":"Negative"\}

18. \{"Food policy":"Negative", "Food quality":"Negative"\}

19. \{"Volunteer appreciation":"Negative"\}

20. \{"Inclusivity":"Negative", "Accessibility":"Negative", "Food diversity":"Negative"\}

\end{tcolorbox}

\begin{tcolorbox}[colback=white, colframe=black, title=Model Response (Gemini)]
1. \{"organization": "Negative", "direction": "Negative", "help": "Negative"\}

2. \{"weather": "Negative", "forecast": "Negative"\}

3. \{"sign-up process": "Negative", "website": "Negative", "IT": "Negative"\}

4. \{"food": "Negative", "options": "Negative", "overpriced": "Negative"\}

5. \{"parking situation": "Negative", "info": "Negative"\}

6. \{"Wi-Fi": "Negative"\}

7. \{"preparation": "Negative", "info": "Negative"\}

8. \{"weather": "Negative", "shade": "Negative", "sunburn": "Negative"\}

9. \{"equipment": "Negative"\}

10. \{"organization": "Negative", "vibe": "Negative"\}

11. \{"schedule": "Negative"\}

12. \{"water": "Negative", "irresponsible": "Negative"\}

13. \{"communication": "Negative", "announcements": "Negative"\} 

14. \{"activities": "Negative", "canceled": "Negative"\}

15. \{"disorganization": "Negative", "stressful": "Negative"\}

16. \{"volunteer shirts": "Negative"\}

17. \{"medical tent": "Negative", "unprepared": "Negative"\}

18. \{"food policy": "Negative", "their food": "Negative"\}

19. \{"appreciation for volunteers": "Negative"\}

20. \{"inclusivity": "Negative", "accessibility": "Negative", "diversity in food choices": "Negative"\} 

\end{tcolorbox}

\newpage

\section{Instructions for Volunteers}
\label{sec:volinstr}

You will be provided with feedback and two sets of annotations. Your task is to choose the annotation set that best captures the feedback based on the following criteria:

\begin{enumerate}
    \item Assess whether the set clearly identifies \textbf{most of}  the relevant aspects \textit{without introducing irrelevant or redundant ones}. 
    \item Assess if the sentiment (positive, negative, or neutral) attached to each aspect correctly reflects the feedback’s tone.

\end{enumerate}

\subsection*{Instructions for Selecting the Set:}
\begin{itemize}
    \item If you prefer \textbf{Set 1}, mark your choice as \textbf{1}.
    \item If you prefer \textbf{Set 2}, mark your choice as \textbf{2}.
    \item If \textbf{both} sets represent the same aspects and associated polarities, mark your choice as \textbf{0}.
\end{itemize}

\subsection*{Example 1:}

\textbf{Feedback}: 
``The food stalls, despite some limitations, did a fantastic job of offering a taste of home to many attendees. It was a nice touch that added to the overall welcoming atmosphere of the event''.

\textbf{Annotation Sets}:
\begin{itemize}
    \item Set 1: \{`Food stalls': `Positive'\}
    \item Set 2: \{`food stalls': `Positive', `atmosphere': `Positive'\}
\end{itemize}

\textbf{Analysis}:
\begin{itemize}
    \item \textbf{Aspects}: 
    \begin{itemize}
        \item Set 1 captures `Food stalls', which is one valid aspect, but it misses the other key aspect, `atmosphere'.
        \item Set 2 captures both `food stalls' and `atmosphere', both of which are valid aspects.
    \end{itemize}
    
    \item \textbf{Sentiment}: 
    \begin{itemize}
        \item Both sets correctly classify the polarity as positive for the aspects they capture.
    \end{itemize}
    
    \item \textbf{Conclusion}: 
    \begin{itemize}
        \item Set 1 identifies only `Food stalls', which is relevant but misses the additional positive aspect related to `atmosphere', while Set 2 provides a more complete annotation, identifying both 'food stalls' and 'atmosphere', which are relevant to the feedback and add no redundant aspects. Thus, in this case, based on the refined criteria, you would select \textbf{2}
    \end{itemize}
\end{itemize}

\subsection*{Example 2:}

\textbf{Feedback}: 
``The food and beverage situation was disappointing, not only in variety but also in accommodating different cultural preferences. It's a basic aspect that should be given more thought in an event celebrating diversity''.

\textbf{Annotation Sets}:
\begin{itemize}
    \item Set 1: \{`Food and beverage diversity': `Negative'\}
    \item Set 2: \{`food and beverage': `Negative', `variety': `Negative', `cultural preferences': `Negative'\}
\end{itemize}

\textbf{Analysis}:
\begin{itemize}
    \item \textbf{Aspects}: 
    \begin{itemize}
        \item Set 1 captures `Food and beverage diversity', which concisely summarizes the feedback and directly reflects the core complaint.
        \item Set 2 introduces `variety', which feels disconnected from `food and beverage' and may add confusion by not clearly aligning with the broader point. It also includes `cultural preferences', which, although mentioned in the feedback, seems redundant because it is disconnected from the major idea.
    \end{itemize}
    
    \item \textbf{Sentiment}: 
    \begin{itemize}
        \item Both sets correctly identify the sentiment as negative for the aspects they capture.
    \end{itemize}
    
    \item \textbf{Conclusion}: 
    \begin{itemize}
        \item Set 1 offers a concise and relevant summary by capturing `Food and beverage diversity', without introducing any irrelevant or redundant information, while Set 2 introduces additional aspects (`variety' and `cultural preferences') that seem disconnected or redundant, making the annotation less relevant and more complicated. Thus, in this case, based on the refined criteria, you would select \textbf{1}.
    \end{itemize}
\end{itemize}

\newpage

\section{Additional Characteristics of Datasets}
\label{sec:stats}

\begin{table*}[h]
\begin{threeparttable}
\caption{Additional Characteristics of Datasets.} 
\centering
\scriptsize
\setlength{\tabcolsep}{2pt} % Adjust the space between columns
\begin{tabular}{|l|c|c|c|c|c|c|c|c|c|}
\hline
 & Total & Total & Unique & Avg & Total & Total & Total Neutral/ & Total & Avg \\
 & Documents & Aspects & Aspects & Aspects/Doc & Positive & Negative & Conflicting & Sentences & Sentences/Doc \\\hline
SemEval-14-Restaurant & 2625 & 4785 & 1545 & 1.82 & 2871 & 986 & 824 & 2660 & 1.01 \\
SemEval-14-Laptop & 1904 & 2950 & 1194 & 1.55 & 1308 & 964 & 619 & 1932 & 1.01 \\
MAMS & 4797 & 12522 & 2659 & 2.61 & 3780 & 3093 & 5649 & 4841 & 1.01 \\
Twitter & 6940 & 6940 & 117 & 1.00 & 1734 & 1733 & 3473 & 12526 & 1.80 \\
\hline
Composite & 16266 & 27197 & 4880 & 1.67 & 9693 & 6776 & 10565 & 21959 & 1.35 \\
\hline
Sports Feedback (Novel) & 480 & 938 & 491 & 1.95 & 405 & 501 & 32 & 1409 & 2.94 \\
\hline
\end{tabular}
\label{tab:absa_dss_addl}
\begin{tablenotes}
\item[1] The number of sentences was obtained using the \textbf{sent\_tokenize} function of \textbf{nltk} (version 3.8.1).
\end{tablenotes}
\end{threeparttable}
\end{table*}

\newpage
\normalsize
\twocolumn

\section{Empirical Selection of \(\theta\) and Metric Analysis}
\label{sec:metricanalysis}
The proposed metric is parameterized by the value of \(\theta\). To select the optimal value of \(\theta\), we established the criterion of maximizing the number of correct aspect pairings while ensuring minimal incorrect aspect pairings.

To evaluate the validity of our proposed metric with the chosen threshold \(\theta\), we conducted the following analyses on a combined dataset, created by merging the test subset of the Novel dataset and a test portion of the Composite dataset:

\begin{enumerate}%[itemsep=3pt, parsep=0pt, topsep=0pt]
    \item Compile a set \( \mathcal{D} \) consisting of the detected aspect sets from all model variations listed in Table~\ref{tab:detailed-results}. The total number of unique detected aspects across all subsets in \( \mathcal{D} \) is given by \( \left| \bigcup_{D_i \in \mathcal{D}} D_i \right| = 10856 \).

    \item For each detected aspect set \( D_i \) in \( \mathcal{D} \) and its corresponding gold aspect set \( G_i \), compute the set difference:
    \setlength{\abovedisplayskip}{1pt}
    \setlength{\belowdisplayskip}{1pt}
    \begin{align*}
    I_i &= i(D_i, G_i, \theta) \setminus \{(d, g) \mid \\
        &\quad d \in D_i,\, g \in G_i,\, d = g\}.
    \end{align*}
    Then, define \( \mathcal{I} \) as the union of these sets over all \( i \):
    \setlength{\abovedisplayskip}{0pt}
    \setlength{\belowdisplayskip}{0pt}
    \[
    \mathcal{I} = \bigcup_{i} I_i.
    \]
    %\vspace{-1em}
    This results in \( \mathcal{I} \), the set of all aspect pairs identified by our metric \( i(D_i, G_i, \theta) \) across all data, but not captured by a simple case-insensitive intersection.

    \item Manually examine all aspect pairs in \( \mathcal{I} \) to assess their validity in relation to the original documents from which they were derived.

\end{enumerate}

\begin{figure}[htbp] % Adjust placement options as needed
    \centering 
    %\hspace*{-10pt} % Adjust the value as needed to move the figure left
    \includegraphics[width=0.49\textwidth]{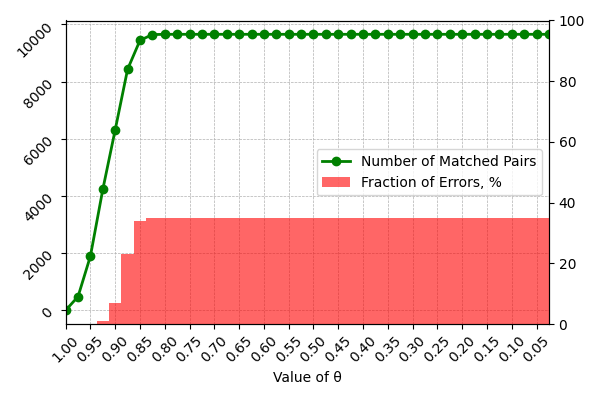}
    \caption{Impact of \(\theta\) on \( | \mathcal{I} | \) and the Fraction of Errors.}

    \label{fig:errpairs} 
\end{figure}

Since \(\theta\) is a real-valued parameter, determining its precise optimal value is infeasible largely due to the requirement for manual analysis of all aspect pairs in \( \mathcal{I} \). Therefore, in this study, we adopt a practical approach by restricting the search space to increments of 0.025 within the interval (0, 1]. Figure~\ref{fig:errpairs} illustrates the effect of \(\theta\) on \( | \mathcal{I} | \) and the fraction of errors introduced by lower \(\theta\) values.

%Appendix~\ref{sec:thetas} 

%\vspace{-0.85em}
%\captionsetup{skip=0pt} 
%\vspace{-1em}

%\color{black}

Manual examination of the pairs in \(\mathcal{I}\) revealed no instances of incorrect aspect pairings for \(\theta = 0.95\), except in 2\% of cases where detected compound aspects were matched with a single gold aspect or vice versa. For example, if \emph{"tomato and onions"} appears as a gold aspect while \emph{"tomato"} and \emph{"onions"} are detected as separate aspects by one of the models, the \(i(D_i, G_i, 0.95)\) approach pairs the gold aspect with \emph{"tomato"}. Despite these exceptions, the proposed metric successfully identified matches not captured by a simple case-insensitive intersection, including the following cases:

\begin{enumerate}%[itemsep=1pt, parsep=0pt, topsep=0pt]
    \item Orthographic Errors: Typographical discrepancies between terms, e.g., \emph{"NBMSG"} and \emph{"NSBG"}, \emph{"atmoshere"} and \emph{"atmosphere"}.
    \item Paraphrastic Variants: Implicit aspects where rearranged word order corresponds to the same concept, such as \emph{"Event variety"} and \emph{"Variety of events"}.
    \item Contextual Elaborations: Aspects identified with additional contextual information, for example, \emph{"Athlete registration"} and \emph{"Athlete registration process"}, \emph{"patties"} and \emph{"full sized patties"}, \emph{"Seagate Momentus XT hybrid drives"} versus \emph{"Two Seagate Momentus XT hybrid drives"}.
    \item Lexical Substitutions: Rephrased aspects demonstrating semantic equivalence, such as \emph{"Food options diversity"} and \emph{"variety of food options"}.
    \item Synonymy: Use of synonyms to express similar concepts, exemplified by \emph{"looks"} and \emph{"appearance"}.
    \item Acronymy: Representation of terms through acronyms, e.g., \emph{"OS"} for \emph{"Operating System"}, \emph{"AC"} for \emph{"Air Conditioning"}.
\end{enumerate}

%\color{red}
Decreasing \(\theta\) to 0.925 introduces a 1\% error rate. These errors primarily stem from terms that are related through a shared context but are not true synonyms. Examples include: \emph{"Alicia Keys"} and \emph{"Aaliyah"}, \emph{"Stephen Colbert"} and \emph{"Jon Stewart"}, \emph{"Barack Obama"} and \emph{"Hillary Clinton"}, \emph{"Xbox"} and \emph{"PlayStation"}, \emph{"Bill Gates"} and \emph{"Microsoft"}, \emph{"iPhone"} and \emph{"WiFi"}, and \emph{"lamb"} and \emph{"chicken"}.

Further decreasing $\theta$ sharply increases the error rate, making it impractical. Thus, we conclude that $\theta = 0.95$ allows the proposed metric to effectively evaluate model performance taking linguistic variation into account while minimizing false-positive pairings.

%Figures~\ref{fig:theta_classic} and \ref{fig:theta_novel} illustrate the effect of \(\theta\) on precision, recall, and \(F_1\) scores for the aspect detection task across individual models and datasets. Specifically, the figures show how changing the values of \(\theta\) impact the 

\onecolumn
\begin{comment}

%subsection{Composite Dataset}
\begin{figure}[htbp]
    \centering
    \includegraphics[width=1\textwidth]{theta_classic.png}
    \caption{Effect of Theta Value on Metrics: Composite Dataset}
    \label{fig:theta_classic}
\end{figure}

\begin{figure}[htbp]
    \centering
    \includegraphics[width=1\textwidth]{theta_novel.png}
    \caption{Effect of Theta Value on Metrics: Novel Dataset}
    \label{fig:theta_novel}
\end{figure}
\end{comment}
\newpage

\section{Experimental Setup}
\label{sec:fthyper}
\begin{table}[h!]
\centering
\caption{Summary of fine-tuning hyperparameters.}
\begin{tabular}{|l|l|}
\hline
\textbf{Hyperparameter}           & \textbf{Value}                \\ \hline
LoRA Attention Dimension (r)       & 128                           \\ \hline
LoRA Alpha                        & 32                            \\ \hline
LoRA Dropout                      & 0.1                           \\ \hline
Bias                              & none                          \\ \hline
Task Type                         & CAUSAL\_LM                    \\ \hline
Per-Device Batch Size              & 8                             \\ \hline
Gradient Accumulation Steps        & 1                             \\ \hline
Learning Rate                     & $1 \times 10^{-4}$            \\ \hline
Optimizer                         & paged\_adamw\_32bit           \\ \hline
Max Training Steps                & varies based on dataset used                           \\ \hline
Warmup Steps                      & 2                             \\ \hline
Mixed Precision (fp16)            & True                          \\ \hline
4-bit Precision                   & True                          \\ \hline
4-bit Double Quantization         & True                          \\ \hline
4-bit Quantization Type           & nf4                           \\ \hline
4-bit Compute Data Type           & bfloat16                      \\ \hline
\textit{Additional Note}      & We saved the model's weights after every 200 steps \\ & and selected the checkpoint just before the \\& validation loss began to increase to avoid overfitting \\ \hline
\end{tabular}

\end{table}

\begin{table}[h!]
\centering
\label{tab:experimental_setup}
\caption{Hardware and Software Used For Experiments.}
\begin{tabular}{|l|l|}
\hline
\textbf{Component} & \textbf{Specification} \\
\hline
\textbf{Hardware} & \\
\hline
GPU              & NVIDIA A100 80GB \\
\hline
CPU              & AMD EPYC 7552 \\
\hline
System Memory    & 128GB DDR4 RAM \\
\hline
\textbf{Software} & \\
\hline
Operating System & Ubuntu 22.04.3 LTS \\
\hline
Python           & 3.10.12 \\
\hline
Transformers     & 4.46.1 \\
\hline
PyTorch          & 2.5.1+cu124 \\
\hline
Datasets & 2.14.7 \\
\hline
bitsandbytes & 0.43.0 \\
\hline
flash-attn & 2.6.3 \\
\hline
PyABSA & 2.3.4 \\
\hline

\end{tabular}

\end{table}

\newpage

\section{Generic ICL}
\label{sec:genericicl}
\tiny
\begin{tcolorbox}[colback=gray!5, colframe=gray!80, title=Generic Prompt]
Given a text, identify aspects and polarities (Positive, Negative, Neutral) as in the examples below. 

Note: the empty dictionary for the third example indicates that there are no aspects or polarities associated with the text:

    TEXT: "I like school but the organization of the art classes needs improvement"
    
    ASPECTS AND POLARITIES: \{"school":"Positive","organization of art classes":"Negative"\}
    
    TEXT: "Dog drinks water"
    
    ASPECTS AND POLARITIES: \{\}
    
    TEXT: "Fall is OK season"
    
    ASPECTS AND POLARITIES: \{"fall":"Neutral"\}
\end{tcolorbox}

\onecolumn
\normalsize
\newpage

\section{Detailed Experimental Results}
\label{sec:expresults}

\begin{table}[ht]
\centering

\caption{Detailed Results for Aspect Extraction Using Adjusted Metrics (\(\theta = 0.95\)) vs Exact Match Evaluation.}

\resizebox{0.63\linewidth}{!}{%
\begin{tabular}{l@{\hspace{15pt}}l@{\hspace{15pt}}l@{\hspace{15pt}}r@{\hspace{10pt}}r@{\hspace{10pt}}r@{\hspace{25pt}}r@{\hspace{10pt}}r@{\hspace{10pt}}r}

\toprule
\textbf{Dataset} & \textbf{Model} & \textbf{Method} & 
\({P}^{.95}\) & \({R}^{.95}\) & \({F_1}^{.95}\) &
\({P}\) & \({R}\) & \({F_1}\) \\
\midrule
% --- Start of rows ---
Laptop-14 & LLaMA-3 & FT Blended & 0.72 & 0.52 & 0.60 & 0.65 & 0.47 & 0.54 \\
Laptop-14 & LLaMA-3 & FT Composite & 0.91 & 0.72 & 0.81 & 0.88 & 0.70 & 0.78 \\
Laptop-14 & LLaMA-3 & FT Novel & 0.43 & 0.42 & 0.42 & 0.34 & 0.33 & 0.34 \\
Laptop-14 & LLaMA-3 & Generic ICL & 0.18 & 0.25 & 0.21 & 0.16 & 0.22 & 0.19 \\
Laptop-14 & LLaMA-3 & ICL with sampling & 0.62 & 0.78 & 0.69 & 0.58 & 0.72 & 0.64 \\
Laptop-14 & Mistral & FT Blended & 0.79 & 0.74 & 0.76 & 0.75 & 0.70 & 0.72 \\
Laptop-14 & Mistral & FT Composite & 0.85 & 0.80 & 0.83 & 0.81 & 0.77 & 0.79 \\
Laptop-14 & Mistral & FT Novel & 0.49 & 0.50 & 0.50 & 0.40 & 0.41 & 0.41 \\
Laptop-14 & Mistral & Generic ICL & 0.33 & 0.60 & 0.43 & 0.25 & 0.46 & 0.33 \\
Laptop-14 & Mistral & ICL with sampling & 0.67 & 0.70 & 0.68 & 0.61 & 0.63 & 0.62 \\
Laptop-14 & PyABSA & - & 0.86 & 0.82 & 0.84 & 0.80 & 0.76 & 0.78 \\
\midrule

Restaurant-14 & LLaMA-3 & FT Blended & 0.72 & 0.58 & 0.64 & 0.66 & 0.54 & 0.59 \\
Restaurant-14 & LLaMA-3 & FT Composite & 0.88 & 0.77 & 0.82 & 0.84 & 0.73 & 0.78 \\
Restaurant-14 & LLaMA-3 & FT Novel & 0.43 & 0.36 & 0.39 & 0.37 & 0.31 & 0.34 \\
Restaurant-14 & LLaMA-3 & Generic ICL & 0.19 & 0.39 & 0.25 & 0.17 & 0.35 & 0.23 \\
Restaurant-14 & LLaMA-3 & ICL with sampling & 0.72 & 0.81 & 0.77 & 0.68 & 0.76 & 0.72 \\
Restaurant-14 & Mistral & FT Blended & 0.80 & 0.85 & 0.83 & 0.75 & 0.80 & 0.77 \\
Restaurant-14 & Mistral & FT Composite & 0.85 & 0.90 & 0.87 & 0.81 & 0.85 & 0.83 \\
Restaurant-14 & Mistral & FT Novel & 0.63 & 0.58 & 0.60 & 0.55 & 0.51 & 0.53 \\
Restaurant-14 & Mistral & Generic ICL & 0.41 & 0.70 & 0.52 & 0.35 & 0.60 & 0.45 \\
Restaurant-14 & Mistral & ICL with sampling & 0.78 & 0.79 & 0.79 & 0.71 & 0.73 & 0.72 \\
Restaurant-14 & PyABSA & - & 0.88 & 0.88 & 0.88 & 0.82 & 0.82 & 0.82 \\
\midrule

Twitter & LLaMA-3 & FT Blended & 0.65 & 0.55 & 0.59 & 0.62 & 0.53 & 0.57 \\
Twitter & LLaMA-3 & FT Composite & 0.96 & 0.86 & 0.91 & 0.96 & 0.85 & 0.90 \\
Twitter & LLaMA-3 & FT Novel & 0.16 & 0.26 & 0.20 & 0.14 & 0.23 & 0.17 \\
Twitter & LLaMA-3 & Generic ICL & 0.13 & 0.33 & 0.19 & 0.12 & 0.29 & 0.17 \\
Twitter & LLaMA-3 & ICL with sampling & 0.75 & 0.85 & 0.80 & 0.70 & 0.79 & 0.74 \\
Twitter & Mistral & FT Blended & 0.80 & 0.83 & 0.81 & 0.79 & 0.82 & 0.80 \\
Twitter & Mistral & FT Composite & 0.92 & 0.92 & 0.92 & 0.91 & 0.91 & 0.91 \\
Twitter & Mistral & FT Novel & 0.21 & 0.37 & 0.27 & 0.18 & 0.33 & 0.24 \\
Twitter & Mistral & Generic ICL & 0.15 & 0.52 & 0.23 & 0.12 & 0.44 & 0.19 \\
Twitter & Mistral & ICL with sampling & 0.65 & 0.80 & 0.72 & 0.60 & 0.74 & 0.66 \\
Twitter & PyABSA & - & 0.43 & 0.32 & 0.37 & 0.32 & 0.23 & 0.27 \\
\midrule

MAMS & LLaMA-3 & FT Blended & 0.67 & 0.46 & 0.54 & 0.62 & 0.42 & 0.50 \\
MAMS & LLaMA-3 & FT Composite & 0.74 & 0.69 & 0.71 & 0.69 & 0.65 & 0.67 \\
MAMS & LLaMA-3 & FT Novel & 0.32 & 0.22 & 0.26 & 0.24 & 0.16 & 0.19 \\
MAMS & LLaMA-3 & Generic ICL & 0.19 & 0.32 & 0.24 & 0.16 & 0.28 & 0.21 \\
MAMS & LLaMA-3 & ICL with sampling & 0.54 & 0.62 & 0.58 & 0.50 & 0.57 & 0.53 \\
MAMS & Mistral & FT Blended & 0.68 & 0.72 & 0.70 & 0.63 & 0.67 & 0.65 \\
MAMS & Mistral & FT Composite & 0.71 & 0.76 & 0.73 & 0.67 & 0.71 & 0.69 \\
MAMS & Mistral & FT Novel & 0.41 & 0.27 & 0.33 & 0.32 & 0.22 & 0.26 \\
MAMS & Mistral & Generic ICL & 0.33 & 0.46 & 0.39 & 0.27 & 0.37 & 0.31 \\
MAMS & Mistral & ICL with sampling & 0.50 & 0.51 & 0.50 & 0.44 & 0.45 & 0.45 \\
MAMS & PyABSA & - & 0.77 & 0.83 & 0.80 & 0.74 & 0.80 & 0.77 \\
\midrule

Novel (ABSA-Sport) & LLaMA-3 & FT Blended & 0.52 & 0.54 & 0.53 & 0.37 & 0.39 & 0.38 \\
Novel (ABSA-Sport)& LLaMA-3 & FT Composite & 0.35 & 0.33 & 0.34 & 0.27 & 0.26 & 0.26 \\
Novel (ABSA-Sport)& LLaMA-3 & FT Novel & 0.54 & 0.54 & 0.54 & 0.37 & 0.38 & 0.37 \\
Novel (ABSA-Sport)& LLaMA-3 & Generic ICL & 0.33 & 0.51 & 0.40 & 0.23 & 0.35 & 0.27 \\
Novel (ABSA-Sport)& LLaMA-3 & ICL with sampling & 0.45 & 0.54 & 0.49 & 0.35 & 0.42 & 0.38 \\
Novel (ABSA-Sport)& Mistral & FT Blended & 0.49 & 0.53 & 0.51 & 0.35 & 0.38 & 0.36 \\
Novel (ABSA-Sport)& Mistral & FT Composite & 0.35 & 0.45 & 0.39 & 0.26 & 0.34 & 0.30 \\
Novel (ABSA-Sport)& Mistral & FT Novel & 0.55 & 0.54 & 0.55 & 0.38 & 0.38 & 0.38 \\
Novel (ABSA-Sport)& Mistral & Generic ICL & 0.21 & 0.44 & 0.29 & 0.14 & 0.30 & 0.19 \\
Novel (ABSA-Sport)& Mistral & ICL with sampling & 0.52 & 0.50 & 0.51 & 0.41 & 0.39 & 0.40 \\
Novel (ABSA-Sport)& PyABSA & - & 0.33 & 0.27 & 0.30 & 0.23 & 0.19 & 0.21 \\
% --- End of rows ---
\bottomrule
\end{tabular}
}

\end{table}

\twocolumn
\normalsize

\end{document}